\pdfoutput=1
\documentclass[11pt]{article}

\usepackage[final]{acl}

\usepackage{makecell}
\usepackage{tabularx}
\usepackage{array}

\usepackage{xcolor}
\usepackage{amsmath}
\usepackage{times}
\usepackage{latexsym}
\usepackage{adjustbox}

\usepackage[most]{tcolorbox}

\usepackage{xcolor}
\usepackage{pifont} % for ding symbols

\usepackage{xspace}

\usepackage{enumitem}

\usepackage{amssymb}   % 提供 \checkmark
\usepackage{booktabs}  % 提供 \toprule, \midrule, \bottomrule
\usepackage{multirow}  % 多行合并

\usepackage{listings}
\usepackage{newunicodechar}

\usepackage[T1]{fontenc}
\usepackage[utf8]{inputenc}

\usepackage{microtype}

\usepackage{inconsolata}

\usepackage{colortbl}
\definecolor{mygray}{gray}{.9}
\definecolor{mypink}{rgb}{.99,.91,.95}
\definecolor{mycyan}{cmyk}{.3,0,0,0}
\usepackage{xcolor}
\definecolor{level1}{RGB}{255, 204, 153} % 浅橙
\definecolor{level2}{RGB}{204, 229, 255} % 浅蓝
\definecolor{level3}{RGB}{204, 255, 204} % 浅绿

\definecolor{hidden-draw}{RGB}{0, 0, 0} % 浅绿

\usepackage{tikz}
\usepackage[edges]{forest}
\usepackage{xcolor}

\definecolor{level0}{rgb}{0.98, 0.68, 0.4}
\definecolor{level1}{rgb}{0.98, 0.92, 0.84}
\definecolor{level2}{rgb}{0.9, 0.9, 0.9}

\definecolor{cat}{HTML}{F4A261}      % 左侧大类（橙色）
\definecolor{subcat}{HTML}{EDE7DB}   % 二级（米色）
\definecolor{leaf}{HTML}{FFFFFF}     % 三级（白色）
\definecolor{outline}{gray}{0.45}

\usepackage{pifont}

\usepackage{graphicx}
\usepackage{forest}
\usepackage{graphicx} % 为 rotatebox 用
\usetikzlibrary{shapes, positioning}

\definecolor{cat}{HTML}{4D7EA8}
\definecolor{subcat}{HTML}{E3EFF7}
\definecolor{leaf}{HTML}{F9FBFD}

\forestset{
  mytree/.style={
    for tree={
      grow'=0,                    % 从左到右
      parent anchor=east,
      child anchor=west,
      anchor=west,
      rounded corners=2pt,
      draw=cat!50,
      edge={draw=cat!50, line width=.5pt},
      align=left,
      inner xsep=4pt,
      inner ysep=3pt,
      l sep=9pt,                  % 兄弟节点间距（竖直）
      s sep=7pt,                  % 父子节点间距（水平）
      font=\footnotesize,        % 默认比正文小一号，适合放在图里
    },
    where level=0{
      fill=cat,
      draw=cat!80,
      text=white,
      font=\bfseries\footnotesize,
      inner xsep=3pt,
      inner ysep=3pt,
      text width=6mm,            % 竖条宽度
      align=center,
    }{},
    where level=1{
      fill=subcat,
      draw=cat!40,
      font=\bfseries\footnotesize,
      text width=28mm,           % 左侧短一些的分类框
    }{},
    where level>=2{
      fill=leaf,
      draw=cat!20,
      text width=90mm,           % 右侧长文本宽度（可以按需要调）
    }{},
  },
}

\usepackage{graphicx}

\title{Cognitive Alpha Mining via LLM-Driven Code-Based Evolution}

\author{%
    \textbf{
    Fengyuan Liu$^{2,3}$\thanks{Corresponding author} \quad
    Yi Huang$^{1}$ \quad
    Sichun Luo$^{2,3}$ \quad
    Yuqi Wang$^{2,3}$ \quad
    Yazheng Yang$^{2,3}$
    }\\
    \textbf{
    Xinye Li$^{2,3}$ \quad
    Zefa Hu$^{1}$ \quad
    Junlan Feng$^{1}$ \quad
    Qi Liu$^{2,3}$\footnotemark[1]
    }\\[3mm]
    $^{1}$Jiutian Research, China Mobile \\
    $^{2}$School of Computing and Data Science, The University of Hong Kong \\
    $^{3}$Grace Investment Machine \\
    \small{\textbf{Correspondence:} \texttt{oxfengyuan@gmail.com}, \texttt{liuqi@cs.hku.hk}}
}

\begin{document}

\definecolor{level1}{RGB}{230,240,255}
\definecolor{level2}{RGB}{210,230,250}
\definecolor{level3}{RGB}{190,220,245}
\definecolor{level4}{RGB}{170,210,240}
\definecolor{level5}{RGB}{150,200,235}
\definecolor{level6}{RGB}{130,190,230}
\definecolor{level7}{RGB}{110,180,225}

\maketitle

\begin{abstract}
Discovering effective predictive signals, or ``alphas,'' from financial data with high dimensionality and extremely low signal-to-noise ratio remains a difficult open problem. Despite progress in deep learning, genetic programming, and, more recently, large language model (LLM)–based factor generation, existing approaches still explore only a narrow region of the vast alpha search space. Neural models tend to produce opaque and fragile patterns, while symbolic or formula-based methods often yield redundant or economically ungrounded expressions that generalize poorly. Although different in form, these paradigms share a key limitation: none can conduct broad, structured, and human-like exploration that balances logical consistency with creative leaps.
To address this gap, we introduce the \emph{Cognitive Alpha Mining Framework (CogAlpha)}, which combines code-level alpha representation with LLM-driven reasoning and evolutionary search. Treating LLMs as adaptive cognitive agents, our framework iteratively refines, mutates, and recombines alpha candidates through multi-stage prompts and financial feedback. This synergistic design enables deeper thinking, richer structural diversity, and economically interpretable alpha discovery, while greatly expanding the effective search space.
Experiments on 5 stock datasets from 3 stock markets demonstrate that CogAlpha consistently discovers alphas with superior predictive accuracy, robustness, and generalization over existing methods. Our results highlight the promise of aligning evolutionary optimization with LLM-based reasoning for automated and explainable alpha discovery.
\end{abstract}

% \vspace{-0.5cm}
\section{Introduction}
% \vspace{-0.2cm}
Alpha mining is the process of discovering predictive financial signals, or ``alphas,'' from financial markets such as the stock market to forecast future asset returns. 
However, since financial markets are characterized by high dimensionality, time-varying volatility~\cite{engle1982autoregressive}, and a low signal-to-noise ratio, it remains challenging to identify explainable, reliable, and diverse alphas that support sustainable profitability and effective risk management. 
Over the decades, alpha mining has undergone several major transformations: from manual construction, to machine learning–driven automation, and more recently, to generative and reasoning-based exploration using LLMs~\cite{guo2024quant}.

In the earliest stage, alpha factors were manually designed by financial experts, grounded in economic intuition and empirical observation. 
Classic examples include the Fama–French factors~\cite{fama1992cross} and various documented financial anomalies~\cite{harvey2016and,hou2017replicating}. 
These human-crafted alphas are interpretable and theoretically sound. However, the design process is inherently labor-intensive and inefficient. 
As financial markets became increasingly complex and data-rich, manual approaches struggled to scale, resulting in diminishing returns and crowding among similar strategies.

To enhance efficiency, researchers began leveraging machine learning models for alpha discovery. 
Some studies directly employed neural networks~\cite{duan2022factorvae,xu2021hist,xu2021rest} to implicitly extract complex and nonlinear alpha structures from market data through deep learning. 
These neural approaches demonstrate strong predictive power and the ability to capture high-dimensional and nonlinear dependencies. 
However, they also suffer from inherent weaknesses: such models often behave as black boxes, making it difficult to trace the underlying decision logic or assess their robustness under changing market conditions. 
As a result, their performance tends to degrade when exposed to regime shifts or unseen patterns.
In contrast, formula-based approaches~\cite{zhang2020autoalpha,zhang2023openfe} aim to identify alphas represented by explicit mathematical expressions. 
Many methods based on genetic programming (GP)~\cite{cui2021alphaevolve,lin2019stock,patil2023ai,schmidt2010age,zhaofan2022genetic} and reinforcement learning (RL)~\cite{liu2021finrl,yu2023generating,shi2025alphaforge} frameworks have been proposed to automatically search symbolic formula spaces. 
These methods provide transparent expressions that are easy to reproduce and evaluate. 
Nonetheless, the resulting formulas are often overly complex or redundant and frequently lack solid economic or financial rationale, causing weak generalization and limited stability in real trading environments.
Despite their differences, both neural and formula-based paradigms share a common limitation: their search processes are inefficient and narrow in scope. 
Neither can emulate human-like reasoning that combines logical consistency with leap-style creativity, leaving a critical gap between algorithmic exploration and genuine conceptual innovation.

Recently, LLMs~\cite{li2025r} have been introduced into alpha mining due to their knowledge integration, abstraction, and generative reasoning capabilities. 
LLMs can synthesize financial knowledge and propose novel formulaic representations at scale. 
Nevertheless, most existing LLM-based approaches~\cite{shi2025navigating,tang2025alphaagent} still rely on formula stacking and pattern repetition rather than genuine reasoning or structural innovation. 
As a result, the generated factors tend to be redundant and susceptible to crowding effects, which limits their sustainability in dynamic market environments. 
The key research gap lies in how to evolve LLMs from mere \emph{pattern replicators} into genuine \emph{cognitive thinkers}. 
Specifically, there remains an unmet need for frameworks that enable LLMs to perform deeper thinking, richer structural diversity, and economically grounded exploration, thereby improving the long-term stability and robustness of the discovered alpha factors. 
Achieving this would move the field beyond brute-force search or shallow formula generation toward a more knowledge-driven and explainable paradigm for alpha discovery.

To bridge this gap, we propose a novel framework named \textbf{CogAlpha} (\emph{Cognitive Alpha Mining}). The name highlights two key aspects of our approach: \emph{Cognitive} and \emph{Alpha}. The term \emph{Cognitive} refers to leveraging iterative feedback from prior generations and agents to enable adaptive generation, thereby moving beyond shallow pattern recognition toward human-like analytical reasoning. The term \emph{Alpha} corresponds to the central goal of discovering profitable signals in quantitative finance. 
By integrating an evolutionary search process that induces deeper thinking in LLMs, together with a seven-level agent hierarchy and a multi-agent quality checker, \textsc{CogAlpha} naturally embodies our vision of advancing toward Cognitive Alpha Mining.

The remainder of this paper is organized as follows. Section~\ref{Sec: Related} reviews related work on LLM-driven alpha mining and deeper LLM thinking. Section~\ref{Sec: Approach} presents the proposed CogAlpha framework in detail, highlighting its seven-level agent hierarchy, multi-agent quality checker, and thinking evolution components. Section~\ref{Sec: Experiments} states the experimental setting and reports experimental results on five different stock datasets, with a primary focus on the CSI300, demonstrating the superiority of our approach. Section~\ref{Sec: Conclusion} concludes the paper and outlines promising directions for future research.

The main contributions of this work are summarized as follows:
\begin{itemize}
    \item We introduce the concept of \textit{Cognitive Alpha Mining}, which opens a new direction for automated, robust, and explainable alpha discovery, and we formalize it through the proposed \textsc{CogAlpha} framework.
    \item We propose a novel method, \textsc{CogAlpha}, which leverages an evolutionary search process that induces deeper thinking in LLMs, together with \textit{a Seven-Level Agent Hierarchy} and \textit{a Multi-Agent Quality Checker}.
    \item Extensive experiments on five datasets from 3 stock markets demonstrate the effectiveness of \textsc{CogAlpha} framework. The alphas extracted by our method exhibit stronger predictive performance, greater stability, and improved interpretability compared with baselines.
\end{itemize}

% \vspace{-0.4cm}
\section{Related Work}
% \vspace{-0.2cm}
\label{Sec: Related}
\paragraph{Alpha Mining with LLM}
Alpha mining is a fundamental task in quantitative finance, aimed at discovering predictive signals, i.e., alpha factors, for stock markets. 
Previous approaches have primarily relied on human experts~\cite{fama1992cross}, genetic programming (GP)~\cite{cui2021alphaevolve,lin2019stock,patil2023ai,schmidt2010age,zhaofan2022genetic}, reinforcement learning (RL)~\cite{liu2021finrl,yu2023generating,shi2025alphaforge}, or deep learning~\cite{duan2022factorvae,xu2021hist,xu2021rest} to explore the vast factor space. 
However, these methods all have inherent limitations: they may be inefficient, produce overly complex solutions, or suffer from limited interpretability. 

Recently, LLMs, with their extensive world knowledge and strong reasoning capabilities, have been introduced into alpha mining. 
For example, AutoAlpha~\cite{kou2024automate} employs LLMs to evaluate and select superior alpha candidates, and agentic frameworks have also been incorporated to enhance adaptivity and automation.
AlphaAgent~\cite{tang2025alphaagent} introduces an agent-based architecture with regularization strategies to mine decay-resistant alpha factors, 
AlphaJungle~\cite{shi2025navigating} presents an LLM-powered Monte Carlo Tree Search (MCTS) framework in which the LLM performs multi-step formula refinement, 
and RD-Agent(Q)~\cite{li2025r} proposes a data-centric feedback loop with factor–model co-optimization that enables continuous factor adaptation under dynamic market conditions. 
However, despite these advances, existing LLM-based alpha mining methods still rely on formulaic search representations, which restrict exploration to shallow regions of the factor space and fail to fully align with LLMs’ strengths in reasoning and code generation. 
In contrast, we leverage LLMs to directly perform code-based evolution, enabling exploration of a broader and deeper search space.

\paragraph{Evolving LLM Thinking}
To further explore the potential of large language models (LLMs), numerous methods have been proposed to enhance their thinking and reasoning capabilities. 
Recent studies have investigated integrating genetic and evolutionary algorithms (EAs) with LLMs. 
For example, Mind Evolution~\cite{lee2025evolving} employs an evolutionary search strategy to scale inference-time computation in large language models. 
WizardLM~\cite{xu2024wizardlm} enhances LLM performance by automatically generating large volumes of open-domain instructions across diverse topics and difficulty levels. 
EvoPrompt~\cite{guoconnecting} combines evolutionary algorithms with LLMs to optimize prompts using operators such as initialization, selection, crossover, mutation, and evaluation, all guided by an LLM; this approach outperforms both human-designed and traditional automated prompts. 
FunSearch~\cite{romera2024mathematical} applies an LLM-guided evolutionary search to discover mathematical heuristics, excelling in constructing novel mathematical objects and advancing algorithmic discovery. 
AlphaEvolve~\cite{novikov2025alphaevolve} further scales this idea by introducing an autonomous evolutionary coding pipeline, where LLMs generate code variants and evaluators iteratively assess and refine them. 
Beyond these studies, evolutionary approaches combined with LLMs have also been explored in text generation~\cite{xiao2023enhancing,jobanputra2025llm} and code generation~\cite{pinna2024enhancing,hemberg2024evolving}. 
Despite these advances, none of the existing works specifically focus on extracting effective signals from highly volatile financial markets.
To this end, we propose \textsc{CogAlpha}, which leverages an evolutionary search process that induces deeper thinking in LLMs, in collaboration with a seven-level agent hierarchy and a multi-agent quality checker, to generate robust and interpretable alpha factors.

% \vspace{-0.2cm}
\section{Approach}
% \vspace{-0.2cm}
\label{Sec: Approach}
The Cognitive Alpha Mining Framework (\textsc{CogAlpha}) is designed to simulate human-like reasoning and discover more sophisticated, logical, and interpretable alpha solutions. 
It employs an evolutionary search strategy that induces deeper thinking in LLMs, together with a seven-level agent hierarchy and a multi-agent quality checker, to perform alpha mining. Each alpha produced by \textbf{CogAlpha} is accompanied by detailed comments that explain its logic, clarify its underlying idea, and present the corresponding formula. Following the comments, the implementation code is provided.
In this section, we introduce the core components of \textsc{CogAlpha} and explain how each part functions within the overall framework.

\vspace{-0.1cm}
\subsection{Seven-Level Agent Hierarchy}
\vspace{-0.1cm}
The only raw factors available are \textit{open}, \textit{high}, \textit{low}, \textit{close}, and \textit{volume} (OHLCV). 
Based on the five factors, we design a seven-level agent hierarchy to explore alphas as comprehensively as possible. 
This hierarchy consists of 21 unique agents. 
As shown in Figure~\ref{fig: SevelLevelAgents}, from a macroscopic (Level~I) to a microscopic (Level~VII) perspective, these agents are organized into seven hierarchical levels. 
Each agent is dedicated to exploring a distinct alpha-discovery direction and independently generates a set of alpha factors according to its designated exploration strategy. 
The following provides a brief overview of each level's exploration domain, and more details are provided in Appendix~\ref{A: Seven-Level Agent Hierarchy}.

% \begin{itemize}[leftmargin=0.5em, label={}]
    % \item \textbf{Level~I: Market Structure \& Cycle Layer} 
    \paragraph{Level~I: Market Structure \& Cycle Layer}
    (\textit{AgentMarketCycle}, \textit{AgentVolatilityRegime}) — Explores large-scale temporal structures such as long-term trends, market phases, and cyclical state transitions inferred from daily OHLCV dynamics.

    % \item \textbf{Level~II: Extreme Risk \& Fragility Layer}
    \paragraph{Level~II: Extreme Risk \& Fragility Layer}
    (\textit{AgentTailRisk}, \textit{AgentCrashPredictor}) — Models tail-risk exposure, crash precursors, and systemic fragility patterns that indicate potential regime breakdowns or stress accumulation.

    % \item \textbf{Level~III: Price–Volume Dynamics Layer} 
    \paragraph{Level~III: Price–Volume Dynamics Layer}
    (\textit{AgentLiquidity}, \textit{AgentOrderImbalance}, \textit{AgentPriceVolumeCoherence}, \textit{AgentVolumeStructure}) — Captures the interaction between price and trading activity---liquidity, order imbalance, and coherence between price movement and volume behavior.

    % \item \textbf{Level~IV: Price–Volatility Behavior Layer} 
    \paragraph{Level~IV: Price–Volatility Behavior Layer}
    (\textit{AgentDailyTrend}, \textit{AgentReversal}, \textit{AgentRangeVol}, \textit{AgentLagResponse}, \textit{AgentVolAsymmetry}) — Analyzes trend persistence, short-term reversal, volatility clustering, and asymmetric price dynamics as the core source of predictive alpha.

    % \item \textbf{Level~V: Multi-Scale Complexity Layer} 
    \paragraph{Level~V: Multi-Scale Complexity Layer}
    (\textit{AgentDrawdown}, \textit{AgentFractal}) — Measures cross-scale irregularity, fractal roughness, drawdown--recovery geometry, and long-memory characteristics in time-series structure.

    % \item \textbf{Level~VI: Stability \& Regime-Gating Layer} 
    \paragraph{Level~VI: Stability \& Regime-Gating Layer}
    (\textit{AgentRegimeGating}, \textit{AgentStability}) — Assesses temporal stability and constructs adaptive gating mechanisms that regulate signal activation under varying market conditions.

    % \item \textbf{Level~VII: Geometric \& Fusion Layer}
    \paragraph{Level~VII: Geometric \& Fusion Layer}
    (\textit{AgentBarShape}, \textit{AgentCreative}, \textit{AgentComposite}, \textit{AgentHerding}) — Focuses on geometric pattern representation (candlestick morphology) and multi-factor fusion, combining independent signals into coherent composites.
% \end{itemize}

\begin{figure*}[t] 
    \centering
    \footnotesize
    % \fbox{\rule[-.5cm]{0cm}{4cm} \rule[-.5cm]{4cm}{0cm}}
    \includegraphics[width=0.95\linewidth]{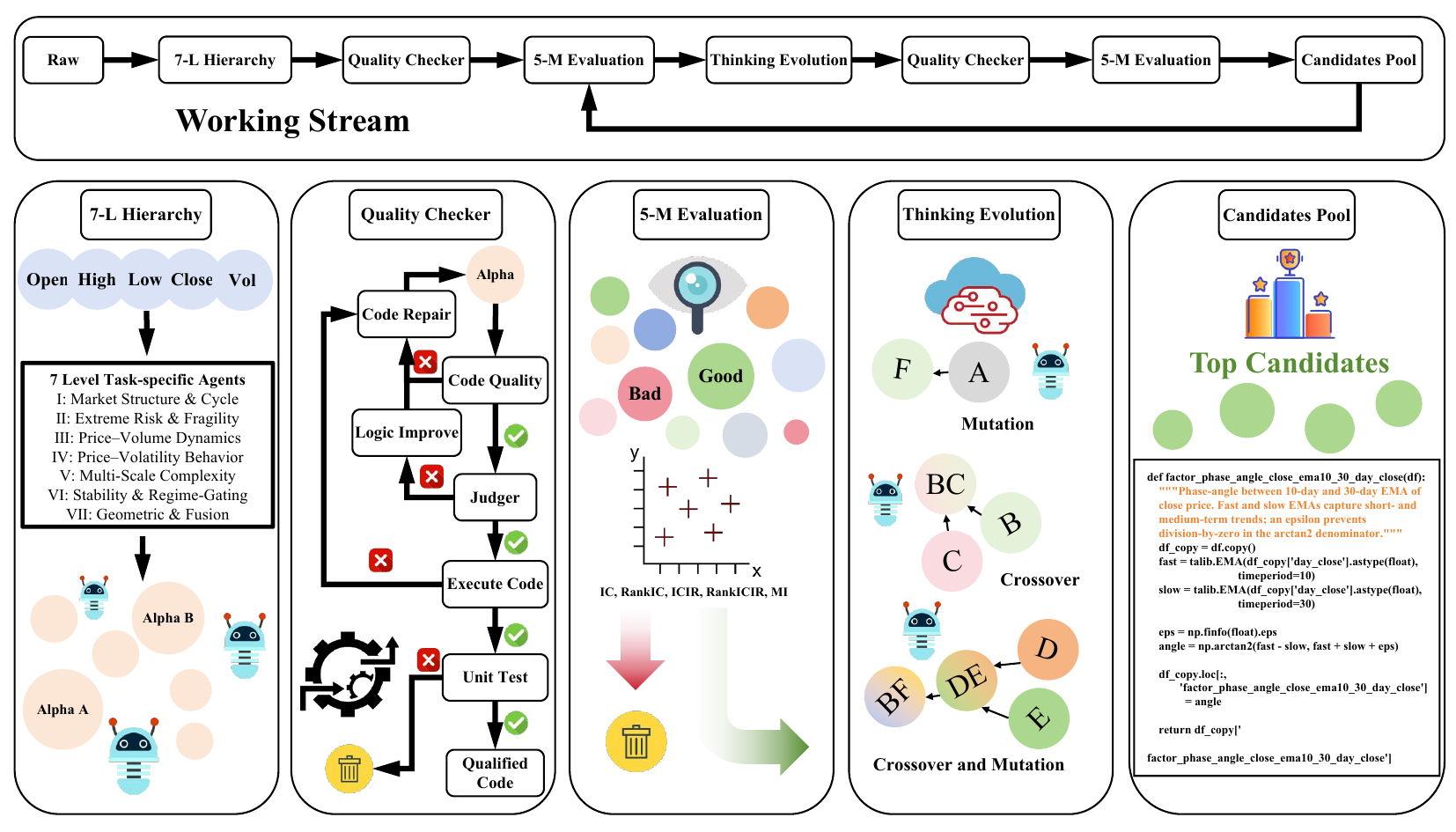}
    \vspace{-0.4cm}
    % \fbox{\rule{0pt}{2.5in} \rule{0.9\linewidth}{0pt}}
    \caption{\textbf{Overview of CogAlpha.} 
    % The figure illustrates how \textsc{CogAlpha} searches for and generates alphas based on OHLCV data. 
    The Seven-Level Agent Hierarchy produces initial alphas derived from the OHLCV data. 
    The Multi-Agent Quality Checker verifies the validity and quality of each generated alpha code. 
    The Filtering module evaluates all alpha codes using five predictive power metrics. 
    Finally, the Thinking Evolution module iteratively refines and recombines qualified candidates through deeper reasoning by LLMs in each iteration.}
    \label{fig: CogAlpha}
    \vspace{-0.5cm}
\end{figure*}

\vspace{-0.1cm}
\subsection{Diversified Guidance}
\vspace{-0.1cm}
To achieve more precise and comprehensive exploration along each alpha-discovery direction, we extend the original guidance generation with five paraphrasing modes: \textit{light}, \textit{moderate}, \textit{creative}, \textit{divergent}, and \textit{concrete}. 
The \textit{light} version performs minimal rewording to maintain almost identical meaning, ensuring linguistic consistency for baseline comparison. 
The \textit{moderate} version introduces natural phrasing variations to enrich expression while keeping the same analytical focus. 
The \textit{creative} version adds interpretative depth and research-oriented nuance to inspire alternative reasoning within the same conceptual boundary. 
The \textit{divergent} version explores new but related analytical perspectives, helping generate complementary hypotheses beyond the original phrasing. 
Finally, the \textit{concrete} version transforms abstract descriptions into measurable, implementation-oriented forms by specifying possible formulas, ratios, or statistical operations. 
Together, these five paraphrasing styles enable broader semantic coverage and deeper factor reasoning without departing from the original analytical intent.
More details about those paraphrasing modes are provided in Appendix~\ref{A: Diversified Guidance}.

% \vspace{-0.2cm}
\subsection{Multi-Agent Quality Checker}
% \vspace{-0.2cm}
To verify the validity and quality of the generated alpha codes, we design a \textit{Multi-Agent Quality Checker}---comprising the \textit{Judge Agent}, \textit{Logic Improvement Agent}, \textit{Code Quality Agent}, and \textit{Code Repair Agent}. 
All alpha codes that pass the quality checker are stored in the candidate pool; otherwise, invalid codes are sent back to the multi-agent system for repair. 
Codes that cannot be repaired or improved after several attempts are discarded. 

As illustrated in Figure~\ref{fig: CogAlpha}, the \textit{Code Quality Agent} first detects issues such as syntax errors, formatting inconsistencies, and runtime bugs. 
If such issues are found, the \textit{Code Repair Agent} attempts to fix the problematic alpha codes based on the feedback provided by the \textit{Code Quality Agent}. 
Next, the \textit{Judge Agent} evaluates whether an alpha factor is logically consistent, technically correct, and economically meaningful. 
If improvement is needed, the \textit{Logic Improvement Agent} refines and enhances alpha codes that fail the \textit{Judge Agent}'s assessment. 
After passing all quality checks, each code is executed. We evaluate numerical stability by detecting runtime errors, the proportion of \textit{NaN} values, overflow/underflow, and distinct values per day. Codes that fail are either rejected or sent back to earlier agents for correction.
If it runs successfully, a unit test is performed to examine potential information leakage. 
Codes that pass the unit test are deemed qualified and stored in the candidate pool. More details are shown in Appendix~\ref{A: Multi-Agent Quality Checker}.

% \vspace{-0.2cm}
\subsection{Fitness Evaluation}
% \vspace{-0.2cm}
After passing the Multi-Agent Quality Checker, each alpha is evaluated using five predictive power metrics: Information Coefficient (\textbf{IC}), Information Coefficient Information Ratio (\textbf{ICIR}), Rank Information Coefficient (\textbf{RankIC}), Rank Information Coefficient Information Ratio (\textbf{RankICIR}), and Mutual Information (\textbf{MI}). 
The first 4 metrics measure the linear relationship between the alpha and the target return, whereas \textbf{MI} captures the nonlinear dependency between them. 
Detailed definitions of these metrics are provided in Appendix~\ref{A: Metrics}.

We set threshold values to identify \textit{qualified} and \textit{elite} alphas. 
Alphas whose five evaluation metrics all exceed the 65th percentile among all alphas in the same generation are classified as \textit{qualified alphas}, 
while those exceeding the 80th percentile are considered \textit{elite alphas}. We constrain each metric by a minimum bound to prevent the dominance of outliers:
\textit{IC} and \textit{RankIC} are bounded below by 0.005, 
\textit{ICIR} and \textit{RankICIR} by 0.05, 
and \textit{MI} by 0.02. 
For elite factors, the minimum bounds are slightly higher: 0.01 for \textit{IC} and \textit{RankIC}, 
0.1 for \textit{ICIR} and \textit{RankICIR}, 
and 0.02 for \textit{MI}. 
The qualified alphas form the new parent pool and are passed to the next iteration, whereas the elite alphas are carried forward and stored in the final candidate pool.
Additionally, the top two elite alphas from the previous generation are carried forward to the next generation to preserve high-quality solutions. 

% \vspace{-0.1cm}
\subsection{Adaptive Generation}
% \vspace{-0.1cm}
After each fitness evaluation, there exists a population of valid alphas and another of invalid alphas, each with different underlying causes. 
To ensure that agents can continuously learn from previous generations, information about both valid and invalid alphas is incorporated into the prompt. 
For each generation, we randomly select two valid alphas and two worst-performing invalid alphas as guiding samples. 
Each selected alpha is first analyzed and summarized to explain why it is valid or invalid. 
Subsequently, the combined fitness results and analytical summaries of the selected alphas are incorporated into the generation prompts, based on which new alphas are generated.

% \vspace{-0.2cm}
\subsection{Thinking Evolution}
% \vspace{-0.2cm}
To guide the LLM to reason more deeply about alpha searching, we employ \textit{Thinking Evolution} to enhance its alpha-mining capability. 
All qualified alphas are evolved through this process. 
As illustrated in Figure~\ref{fig: CogAlpha}, \textit{Thinking Evolution} implements a genetic-style optimization process in natural language space, where candidate alpha codes undergo mutation and crossover operations expressed through textual prompts.
It consists of 2 agents: the \textit{Mutation Agent} and the \textit{Crossover Agent}. 
The \textit{Mutation Agent} slightly modifies a given alpha code to introduce variability, whereas the \textit{Crossover Agent} generates a new alpha code by combining two existing ones. 
Three types of evolution are conducted: mutation only, crossover only, and crossover followed by mutation. 
After each evolution step, the resulting alpha codes are examined by the Multi-Agent Quality Checker. 
This process continues until all generations are completed.
% Without using islands

\vspace{-0.1cm}
\section{Experiments}
\vspace{-0.1cm}
\label{Sec: Experiments}
In this section, we first describe the experimental settings and compare our framework with the baselines. 
Then, we demonstrate the interpretability and evolutionary process of the generated alphas. 
Finally, we study the sensitivity of our method to different metric thresholds.

\subsection{Experimental Settings}
\vspace{-0.1cm}
\paragraph{Datasets} Our experiments are mainly conducted on the CSI300 (China Securities Index 300). Its components consist of 300 large-cap A-share stocks in the Chinese market. We primarily use the 10-day return as the prediction target, with buying and selling at the open price. The dataset is split chronologically into training (2011/01/01-2019/12/31), validation (2020/01/01-2020/12/31), and test (2021/01/01-2024/12/01) periods. Four other datasets (CSI500, S\&P500, HSI, and HSCI) from three different stock markets (China, U.S. and HK) are also tested. More details and results are provided in Appendix~\ref{A: Datasets} and Appendix~\ref{A: Different Settings}.

\paragraph{Model}
In our paper, all agents are based on \textbf{gpt-oss-120b}~\cite{openai2025gptoss} by default. 
For the task-specific agents in the seven-level agent hierarchy and the thinking-evolution agents, the temperature is randomly selected from \{0.7, 0.8, 0.9, 1.0, 1.1, 1.2\} to encourage diversity. 
For the agents in the multi-agent quality checker, the temperature is fixed at 0.8. 
The maximum token length is set to 4096. 
By default, LightGBM~\cite{ke2017lightgbm} is used to train the alphas generated by our method.

\paragraph{Training Setting}
The size of the initial pool is set to 80, meaning that the minimum number of alphas generated by the task-specific agents is 80. 
The parent pool size is set to 32, indicating that after filtering, at most 32 alphas are retained and passed to the next generation. 
The children's pool is set to be three times the size of the parent pool, meaning that the minimum number of alphas generated by the evolutionary agents is 96. 
By default, each task-specific agent leads a complete evolutionary cycle, which consists of 24 generations and 3 inner sub-cycles, with each sub-cycle comprising 8 generations. 
Thus, each task-specific agent initiates the evolutionary search 3 times.  
In addition, every 2 generations, new alphas generated by the task-specific agents are filtered and injected into the parent pool. 
For each generation, the top two elite alphas from the previous generation are always carried forward to the next. 
All alphas containing more than 30\% NaN values or failing the multi-agent quality checker are discarded. 
All experiments are conducted on NVIDIA H100 GPUs.

\paragraph{Evaluation}
We use four predictive power metrics to evaluate the performance of alpha combinations: the Information Coefficient (\textbf{IC}), Information Coefficient Information Ratio (\textbf{ICIR}), Rank Information Coefficient (\textbf{RankIC}), and Rank Information Coefficient Information Ratio (\textbf{RankICIR}). 
The \textbf{IC} measures the linear correlation between alpha values and subsequent total returns, reflecting the overall predictive power of the alpha. 
The \textbf{ICIR} quantifies the stability and temporal consistency of \textbf{IC}. 
The \textbf{RankIC} and \textbf{RankICIR} are similar to \textbf{IC} and \textbf{ICIR}, respectively, but they measure the monotonic relationship between the alpha and subsequent total returns rather than linear correlation. 
In addition, two performance indicators are employed: the Information Ratio (\textbf{IR}) and Annualized Excess Return (\textbf{AER}). 
The \textbf{IR} evaluates the risk-adjusted excess return, and the \textbf{AER} measures the annualized excess cumulative return over a given period.
Detailed definitions and formulas of these metrics are provided in Appendix~\ref{A: Metrics}.

\setlength{\textfloatsep}{10pt}
\begin{table*}[t]
\centering
\scriptsize  
\resizebox{\textwidth}{!}{
\begin{tabular}{llccccccccc}
\toprule
\multicolumn{2}{c}{\multirow{3}{*}{\textbf{Models}}} & 
\multicolumn{1}{c}{\multirow{3}{*}{\textbf{Dataset}}} & 
\multicolumn{1}{c}{\multirow{3}{*}{\textbf{Horizon}}} & 
\multicolumn{6}{c}{\textbf{CSI300}} \\ 
\cmidrule(lr){5-10}
\multicolumn{4}{c}{} & \textbf{IC} & \textbf{RankIC} & \textbf{ICIR} & \textbf{RankICIR} & \textbf{AER} & \textbf{IR} \\ %& \textbf{MDD} \\
\midrule
\multirow{7}{*}{\begin{tabular}[c]{@{}l@{}}Machine-Learning \end{tabular}} 
& Linear & \multicolumn{1}{c}{\multirow{7}{*}{CSI300}} & \multicolumn{1}{c}{\multirow{7}{*}{10 days}} & 0.0165 & 0.0211 & 0.1612 & 0.1655 & -0.0076 & -0.0756 \\
& MLP & & & 0.0227 & 0.0327 & 0.2227 & 0.3037 & 0.0678 & 0.9351 \\ 
& RandomForest & & & 0.0240 & 0.0410 & 0.2932 & \textbf{0.4385} & 0.0784 & 0.8381 \\
& LightGBM & & & 0.0269 & 0.0412 & 0.2811 & 0.3327 & 0.0878 & 1.0980 \\ 
& XGBoost & & & 0.0257 & 0.0376 & 0.2783 & 0.4093 & 0.1081 & 1.3166 \\ 
& CatBoost & & & 0.0197 & 0.0239 & 0.2196 & 0.3043 & 0.0462 & 0.5373 \\ 
& Adaboost & & & 0.0187 & 0.0284 & 0.2709 & 0.3369 & 0.1138 & 1.2633 \\ 
\midrule
\multirow{4}{*}{\begin{tabular}[c]{@{}l@{}}Deep-Learning \end{tabular}} 
& Transformer & \multicolumn{1}{c}{\multirow{4}{*}{CSI300}} & \multicolumn{1}{c}{\multirow{4}{*}{10 days}} & -0.0090 & -0.0022 & -0.0800 & -0.0181 & 0.0492 & 0.6361 \\ 
& GRU & & & 0.0074 & 0.0176 & 0.0747 & 0.1370 & 0.0335 & 0.3386 \\ 
& LSTM & & & 0.0096 & 0.0216 & 0.0886 & 0.1619 & 0.0593 & 0.6030 \\ 
& CNN & & & 0.0268 & 0.0392 & 0.2432 & 0.3117 & 0.0763 & 0.9642 \\ 
\midrule
\multirow{4}{*}{\begin{tabular}[c]{@{}l@{}}Libraries and Methods \end{tabular}} & Alpha $158$ & \multicolumn{1}{c}{\multirow{4}{*}{CSI300}} & \multicolumn{1}{c}{\multirow{4}{*}{10 days}} & 0.0358 & 0.0402 & 0.2737 & 0.2866 & 0.0946 & 0.8556 \\ 
& Alpha $360$ & & & 0.0200 & 0.0136 & 0.1674 & 0.1067 & 0.1198 & 1.0762 \\ 
& AutoAlpha & & & 0.0211 & 0.0177 & 0.2030 & 0.1588 & 0.0658 & 0.6307 \\
% \midrule
& AlphaAgent & & & 0.0246 & 0.0289 & 0.2407 & 0.2721 & 0.1072 & 1.2310 \\ 
% RD-Agent(Q) & gpt-4o & & &  \\ 
\midrule
\multirow{6}{*}{\begin{tabular}[c]{@{}l@{}}LLM \end{tabular}}
& Llama3 8B & \multicolumn{1}{c}{\multirow{6}{*}{CSI300}} & \multicolumn{1}{c}{\multirow{6}{*}{10 days}} & 0.0121 & -0.0074 & 0.0972 & -0.0540 & 0.0520 & 0.5077 \\ 
& Llama3 70B & & & 0.0205 & 0.0229 & 0.1786 & 0.1915 & 0.0681 & 0.6312 \\ 
& gpt-oss-20B & & & 0.0061 & 0.0075 & 0.0613 & 0.0680 & 0.0464 & 0.4885 \\ 
& gpt-oss-120B & & & 0.0300 & 0.0318 & 0.2501 & 0.2595 & 0.0789 & 0.8015 \\ 
& GPT-4.1 & & & 0.0118 & 0.0114 & 0.1069 & 0.1037 & 0.0360 & 0.3628 \\ 
& o3 & & & 0.0019 & -0.0050 & 0.0203 & -0.0475 & 0.0218 & 0.2278 \\ 
\midrule
\multirow{1}{*}{\begin{tabular}[c]{@{}l@{}}CogAlpha \end{tabular}}
& gpt-oss-120B & CSI300 & 10 days & \textbf{0.0591} & \textbf{0.0814} & \textbf{0.3410} & 0.4350 & \textbf{0.1639} & \textbf{1.8999} \\ 
\bottomrule
\end{tabular}
}
\vspace{-0.2cm}
\caption{Performance comparison between \textsc{CogAlpha} and 21 baseline methods on the CSI~300 constituent stock dataset. 
The best performance values for each task are highlighted in \textbf{bold}.}
\label{T: Baselines}
\vspace{-0.4cm}
\end{table*}

\vspace{-0.1cm}
\subsection{Comparison with Baselines}
\vspace{-0.1cm}
We conduct a comprehensive evaluation of \textsc{CogAlpha} by comparing it against 21 benchmark methods from various application domains (see Table~\ref{T: Baselines}). 
First, we select 7 commonly used machine learning models in quantitative finance: Linear Regression, MLP, Random Forest~\cite{breiman2001random}, LightGBM~\cite{ke2017lightgbm}, XGBoost~\cite{chen2016xgboost}, CatBoost~\cite{prokhorenkova2018catboost}, and AdaBoost~\cite{freund1997decision}. 
Next, we include 4 representative deep learning models: GRU~\cite{cho2014learning}, LSTM~\cite{hochreiter1997long}, CNN~\cite{lecun2002gradient}, and Transformer~\cite{vaswani2017attention}. 
We further incorporate two widely used alpha libraries, Alpha-158~\cite{qlib-alpha158} and Alpha-360~\cite{qlib-alpha360}, as baseline factor sets. We also include two closely related automated alpha mining methods, AutoAlpha~\cite{kou2024automate} and AlphaAgent~\cite{tang2025alphaagent}, for comparison.
In addition, six LLMs are evaluated to assess their capacity for alpha mining. Among them, two are closed-source models: GPT-4.1~\cite{openai2025gpt41}, a non-reasoning model, and o3~\cite{openai2025deepresearch}, a reasoning model. The remaining four are open-source models of different scales and origins: Llama3-8B, Llama3-70B~\cite{grattafiori2024llama}, GPT-OSS-20B, and GPT-OSS-120B~\cite{openai2025gptoss}.
For methods and LLMs that generate alpha factors, we evaluate performance using multi-factor combinations constructed from the 20 generated alphas.

As shown in Table~\ref{T: Baselines}, traditional machine learning methods generally exhibit better overall performance than deep learning methods. There is no substantial performance gap between traditional machine learning models and existing alpha libraries or LLM-driven methods that mine formulaic alphas.
Moreover, from the four predictive metrics of \textsc{Alpha158} and \textsc{Alpha360}, we observe that a larger number of alpha factors does not necessarily lead to higher IC or RankIC values. 
For open-source LLMs, larger models generally exhibit stronger alpha-mining capabilities than smaller ones. Surprisingly, the two closed-source models perform poorly, with the reasoning-oriented model achieving the worst performance among all evaluated LLMs.
Overall, \textsc{CogAlpha} consistently outperforms all baseline methods, achieving superior results across all evaluation metrics. The only exception is the Random Forest model, which may be attributed to the fact that the signals generated by Random Forest exhibit highly stable RankIC values with very low standard deviation, leading to a higher RankICIR.

\subsection{Ablation Study}
In this section, we evaluate the effectiveness of each component of CogAlapha: Adaptive Generation (\textbf{A}), Diversified Guidance (\textbf{G}), Seven-Level Agent Hierarchy (\textbf{H}), and Thinking Evolution (\textbf{E}). As shown in Table~\ref{T: Ablation} in Appendix~\ref{A: Ablation Study}, the four parts can, to some extent, improve the effectiveness and performance of alpha mining.

\subsection{Interpretability of Generated Alpha}
In this section, we analyze the interpretability of the generated alphas. Each alpha produced by \textbf{CogAlpha} is accompanied by detailed comments that explain its logic, clarify its underlying idea, and present the corresponding formula. Following the comments, the implementation code is provided.

The Python code in Listing~\ref{lst: alpha_code1} is an example of a generated alpha. It measures the \textbf{liquidity impact}: the price rise (high -- close) per unit of traded volume. 
\begin{equation}
\mathrm{Alpha} = \frac{day_{high} - day_{close}}{day_{volume} + \varepsilon}.
\end{equation}
A large positive value indicates that the stock price increased sharply while trading volume remained low, implying thin liquidity and a higher expected short-term return. This design can be interpreted as a measure of the \textit{price impact per unit of traded volume}. In market microstructure theory, this reflects the liquidity constraint between price movements and trading volume: large price changes under low volume often signal poor liquidity, an imbalanced order book, and markets where small trades can move prices significantly. 
Such conditions may imply short-term reversal or momentum effects, consistent with the findings of \textit{Continuous Auctions and Insider Trading}~\cite{kyle1985continuous} and \textit{Illiquidity and Stock Returns}~\cite{amihud2002illiquidity} on price impact and illiquidity-return relationships.
\begin{lstlisting}[caption={Initial alpha measuring liquidity impact}, label={lst: alpha_code1}, basicstyle=\tiny]
def factor_upward_impact_per_vol(df):
    """Liquidity-impact: price rise (high-close) per unit of traded volume. A large positive value means the stock moved up strongly while volume stayed low, indicating thin liquidity and higher expected short-term return. 
    Formula: (high – close) / (volume + ε)."""
    df_copy = df.copy()
    eps = 1e-9
    df_copy['price_up'] = df_copy['high'] - df_copy['close']
    df_copy['factor_upward_impact_per_vol'] = df_copy['price_up'] / (df_copy['volume'] + eps)
    return df_copy['factor_upward_impact_per_vol']
\end{lstlisting}

\subsection{Evolution of Alphas}
To demonstrate the evolution capability of \textbf{CogAlpha}, we show an example of how liquidity-related alphas evolve over multiple iterations. Each generated alpha is evaluated by predictive metrics (IC and RankIC). Poorly performing alphas are automatically filtered out, while stronger ones are preserved and further evolved.

The first version (\autoref{lst: alpha_code1}) represents an initial manually designed alpha. It measures the liquidity impact as the price rise \((high - close)\) per unit of traded volume. Its metrics are: \textbf{IC: 0.0090}, \textbf{RankIC: 0.0061}. Through mutation, the model generates an alternative formulation (\autoref{lst: alpha_code2} in Appendix~\ref{A: Evolution of Alphas}) that uses the full daily price range \((high - low)\) instead of the closing difference. This captures broader intraday liquidity behavior. Its metrics slightly decrease to \textbf{IC: 0.0073}, \textbf{RankIC: 0.0021}, and thus this version is discarded in later rounds. After several evolutionary rounds, CogAlpha produces a more refined version (\autoref{lst: alpha_code3} in Appendix~\ref{A: Evolution of Alphas}). It normalizes the absolute daily price move by dollar volume and applies a $\tanh$ transformation to ensure boundedness and robustness. The evolved alpha achieves significantly improved performance with \textbf{IC: 0.0141} and \textbf{RankIC: 0.0087}, showing the ability of the evolutionary mechanism to refine quantitative factors effectively.

After a complete evolution cycle, CogAlpha is able to generate a large number of single-factor alphas with strong predictive power, many of which achieve \textbf{absolute IC values above 0.05} and \textbf{absolute RankIC values above 0.07}. This demonstrates the framework's capacity to autonomously explore and optimize factor space toward higher-performing and more interpretable alphas.

\subsection{Generalization to Different Settings}
To test the generalization of CogAlpha, we conduct experiments on five different datasets (CSI300, CSI500, S\&P500, HSI, HSCI) from three different stock markets (China, U.S., and HK), using two training methods (LightGBM and Ridge) and two prediction horizons (10 days and 30 days). As shown in Table~\ref{T: Different Settings} in Appendix~\ref{A: Different Settings}, our method consistently performs well across different settings.

\subsection{Different Fitness Threshold} 
In this section, we analyze the sensitivity of our method to different threshold settings used for filtering alpha factors. 
To maintain the quality of the selected alphas, we experiment with three threshold pairs: (65, 80), (80, 90), and (85, 95). 
In each pair, the former value represents the percentile threshold for \textit{qualified factors} that advance to the next generation, while the latter corresponds to the percentile threshold for \textit{elite factors} that are directly stored in the final candidate pool. 
For comparison, we also establish a baseline configuration to ensure the quality consistency of filtered alphas. 
Specifically, thresholds for each predictive metric are determined based on the empirical distribution of factor scores. 
We constrain each metric by a minimum bound to prevent the dominance of outliers:
\textit{IC} and \textit{RankIC} are bounded below by 0.005, 
\textit{ICIR} and \textit{RankICIR} by 0.05, 
and \textit{MI} by 0.02. 
For elite factors, the minimum bounds are slightly higher: 0.01 for \textit{IC} and \textit{RankIC}, 
0.1 for \textit{ICIR} and \textit{RankICIR}, 
and 0.02 for \textit{MI}. 
As shown in Figure~\ref{fig: Cumulative Return Different Threshold} in Appendix~\ref{A: Different Fitness Threshold}, the threshold pair (65, 80) yields better overall performance. 
This result may be attributed to the larger parent pool size under this configuration, which encourages evolutionary search to explore a broader alpha space and mitigates the risk of premature convergence to local optima.

\vspace{-0.2cm}
\section{Conclusion}
\vspace{-0.2cm}
\label{Sec: Conclusion}
In this work, we study how to extract interpretable and reliable alpha signals from financial markets characterized by high volatility and a low signal-to-noise ratio. We introduce the concept of \textit{Cognitive Alpha Mining}, which opens a new direction for automated, robust, and explainable alpha discovery. We further propose \textsc{CogAlpha}, a multi-agent framework powered by deeper-thinking LLMs. Extensive experiments demonstrate the effectiveness of our approach. In future work, we plan to implement our method in live trading environments to further validate its practical performance.

\section*{Limitations}
The CogAlpha framework is intended for academic use only and does not provide any financial opinions. The backtesting simulations are implemented and executed entirely within the Qlib framework, which may not fully replicate the conditions of live trading environments. Additionally, due to the inherent randomness of LLM outputs, reproducing exactly the same alphas in each run can be challenging. Furthermore, the execution time of the experiments is influenced by the size of the dataset, with larger datasets potentially leading to longer processing times.

\section*{Ethical Considerations}
All datasets used in this paper were downloaded from public sources and are publicly available. 

We used OpenAI's ChatGPT-5.2 for grammar checking and suggestions, but manually verified all edits. 
No AI-generated content was directly included in the final submission.

Users of the CogAlpha framework and its related code are responsible for sourcing their own financial data and independently evaluating the risks associated with the generated factors and models in their specific contexts. It is crucial to approach the agent-generated code, data, and models with care and perform comprehensive verification. The CogAlpha framework does not offer financial advice and is not intended to substitute the expertise of qualified financial professionals in the creation, evaluation, and approval of financial products.

\paragraph{Acknowledgement:} Funded by China Mobile – HKU Joint Innovation Centre (R24113J4, R26110S3)

\bibliography{acl}

\appendix
\section{Approach}
\label{A: Approach}

\subsection{Seven-Level Agent Hierarchy}
\label{A: Seven-Level Agent Hierarchy}

\begin{figure*}[htbp]
\centering
\scalebox{1.0}{
\begin{tikzpicture}[font=\small, align=center]

\node[draw, fill=level1, text width=6cm, minimum height=1cm,
      trapezium, trapezium angle=80] (L1)
    { \textbf{Level I: Market Structure \& Cycle Layer} \\[-1pt]
      \textit{AgentMarketCycle}, \textit{AgentVolatilityRegime} };

\node[below=0cm of L1, draw, fill=level2, text width=6.3cm, minimum height=1cm,
      trapezium, trapezium angle=80] (L2)
    { \textbf{Level II: Extreme Risk \& Fragility Layer} \\[-1pt]
      \textit{AgentTailRisk}, \textit{AgentCrashPredictor} };

\node[below=0cm of L2, draw, fill=level3, text width=7.75cm, minimum height=1cm,
      trapezium, trapezium angle=80] (L3)
    { \textbf{Level III: Price--Volume Dynamics Layer} \\[-1pt]
    \textit{AgentLiquidity}, \textit{AgentOrderImbalance}, \\
    \textit{AgentPriceVolumeCoherence}, \textit{AgentVolumeStructure} };

\node[below=0cm of L3, draw, fill=level4, text width=8.2cm, minimum height=1cm,
      trapezium, trapezium angle=80] (L4)
    { \textbf{Level IV: Price--Volatility Behavior Layer} \\[-1pt]
      \textit{AgentDailyTrend}, \textit{AgentReversal}, \\
      \textit{AgentRangeVol}, \textit{AgentLagResponse}, \textit{AgentVolAsymmetry} };

\node[below=0cm of L4, draw, fill=level5, text width=7.34cm, minimum height=1cm,
      trapezium, trapezium angle=80] (L5)
    { \textbf{Level V: Multi-Scale Complexity Layer} \\[-1pt]
      \textit{AgentDrawdown}, \textit{AgentFractal} };

\node[below=0cm of L5, draw, fill=level6, text width=7.65cm, minimum height=1cm,
      trapezium, trapezium angle=80] (L6)
    { \textbf{Level VI: Stability \& Regime-Gating Layer} \\[-1pt]
      \textit{AgentRegimeGating}, \textit{AgentStability} };

\node[below=0cm of L6, draw, fill=level7, text width=9.34cm, minimum height=1cm,
      trapezium, trapezium angle=80] (L7)
    { \textbf{Level VII: Geometric \& Fusion Layer} \\[-1pt]
      \textit{AgentBarShape}, \textit{AgentCreative}, \\
      \textit{AgentComposite}, \textit{AgentHerding} };

\node[fit=(L1)(L7), inner sep=0pt] (StackBox) {};

\draw[->, thick]
  ($(StackBox.north west)+(-0.8cm,0)$) -- node[pos=0.4, rotate=90, yshift=6pt, left]{\small Micro $\leftarrow$ Macro}
  ($(StackBox.south west)+(-0.8cm,0)$);

\end{tikzpicture}
}
\caption{\textbf{Seven-Level Agent Hierarchy} (Top--Down Pyramid). The pyramid illustrates the seven-level agent hierarchy from macro-structural reasoning to micro-level fusion.}
\label{fig: SevelLevelAgents}
% \vspace{-0.4cm}
\end{figure*}

\begin{table*}[t]
\centering
\renewcommand{\arraystretch}{1.5}
\begin{tabular}{p{1cm} p{5cm} p{8.5cm}}
\toprule
\textbf{Level} & \textbf{Layer Name} & \textbf{Description} \\
\midrule
I & Market Structure \& Cycle Layer & Explores large-scale temporal structures such as long-term trends, market phases, and cyclical state transitions inferred from daily OHLCV dynamics. \\
II & Extreme Risk \& Fragility Layer & Models tail-risk exposure, crash precursors, and systemic fragility patterns that signal potential regime breakdowns or stress accumulation. \\
III & Price–Volume Dynamics Layer & Captures the interactions between price and trading activity—liquidity, order imbalance, and coherence between price movement and volume behavior. \\
IV & Price–Volatility Behavior Layer & Analyzes trend persistence, short-term reversals, volatility clustering, and asymmetric price dynamics as core sources of predictive alpha. \\
V & Multi-Scale Complexity Layer & Measures cross-scale irregularities, fractal roughness, drawdown–recovery geometry, and long-memory characteristics in time-series structures. \\
VI & Stability \& Regime-Gating Layer & Assesses temporal stability and constructs adaptive gating mechanisms that regulate signal activation under varying market conditions. \\
VII & Geometric \& Fusion Layer & Focuses on geometric pattern representation (candlestick morphology) and multi-factor fusion, combining independent signals into coherent composite factors. \\
\bottomrule
\end{tabular}
\caption{Seven-Level Agent Hierarchy of \textsc{CogAlpha} and their corresponding conceptual focuses.}
\label{tab:cogalpha_layers}
\vspace{-0.4cm}
\end{table*}

\begin{itemize}
    \item \textbf{Level 1: Market Structure \& Cycle Layer} \\
    \textit{AgentMarketCycle} explores long-term cyclical transitions and phase shifts in price dynamics, revealing hidden market rhythms and structural turning points.  
    \textit{AgentVolatilityRegime} detects transitions between calm and turbulent volatility states, characterizing regime persistence and clustering behavior.

    \item \textbf{Level 2: Extreme Risk \& Fragility Layer} \\
    \textit{AgentTailRisk} quantifies downside sensitivity and tail-event exposure, modeling how negative shocks propagate through time.  
    \textit{AgentCrashPredictor} identifies early warning signals of market collapses by tracking volatility compression, liquidity depletion, and structural fragility patterns.

    \item \textbf{Level 3: Price–Volume Dynamics Layer} \\
    \textit{AgentLiquidity} measures market depth and trading frictions through price impact and turnover variability.  
    \textit{AgentOrderImbalance} captures directional pressure from one-sided participation inferred from daily OHLCV patterns.  
    \textit{AgentPriceVolumeCoherence} examines synchronization and divergence between price and volume changes, revealing energy alignment or decoupling.  
    \textit{AgentVolumeStructure} analyzes the statistical shape and concentration of trading activity to understand participation rhythm and clustering.

    \item \textbf{Level 4: Price–Volatility Behavior Layer} \\
    \textit{AgentDailyTrend} models directional persistence and multi-day momentum strength to uncover sustained price movements.  
    \textit{AgentReversal} captures mean-reversion and short-term overreaction corrections following transient mispricings.  
    \textit{AgentRangeVol} investigates range-based volatility dynamics, including compression–expansion cycles in daily price ranges.  
    \textit{AgentLagResponse} studies delayed price adjustments and lagged feedback between volatility, volume, and returns.  
    \textit{AgentVolAsymmetry} measures asymmetric volatility between upward and downward price moves, highlighting skewed risk behavior.

    \item \textbf{Level 5: Multi-Scale Complexity Layer} \\
    \textit{AgentDrawdown} evaluates the depth, duration, and recovery geometry of cumulative losses, emphasizing temporal resilience.  
    \textit{AgentFractal} assesses multi-scale roughness and long-memory characteristics through cross-horizon variability and structural irregularity.

    \item \textbf{Level 6: Stability \& Regime-Gating Layer} \\
    \textit{AgentRegimeGating} constructs adaptive gates that modulate signal activation depending on volatility, trend, or liquidity states.  
    \textit{AgentStability} quantifies temporal consistency and persistence in returns or derived signals, emphasizing robustness and smoothness.

    \item \textbf{Level 7: Geometric \& Fusion Layer} \\
    \textit{AgentComposite} fuses multiple independent factors into coherent composites, emphasizing synergy and orthogonality among signals.  
    \textit{AgentCreative} applies non-linear transformations, reparametrizations, or soft gating to generate novel feature representations.  
    \textit{AgentBarShape} encodes candlestick geometry—body, shadow, and symmetry—into continuous and interpretable quantitative descriptors.  
    \textit{AgentHerding} detects collective crowding behavior and directional alignment within OHLCV dynamics, reflecting market consensus intensity.
\end{itemize}

\subsection{Diversified Guidance}
\label{A: Diversified Guidance}
\begin{itemize}
    \item \textbf{Light}: Performs minimal rewording to maintain nearly identical meaning while improving clarity and linguistic fluency. It serves as a baseline for consistency testing across linguistic variations.

    \item \textbf{Moderate}: Rephrases the content naturally with mild enrichment or stylistic variation. This helps capture nuanced semantic differences and tests factor robustness under slightly altered descriptive framing.

    \item \textbf{Creative}: Introduces expressive, research-oriented rewording that adds interpretative depth. This style aims to inspire novel analytical angles or alternative reasoning patterns that remain aligned with the original domain.

    \item \textbf{Divergent}: Produces exploratory rewrites from new but relevant analytical viewpoints, often shifting emphasis toward different sub-mechanisms within the same conceptual framework. This encourages broader hypothesis generation and factor diversity.

    \item \textbf{Concrete}: Makes the guidance more specific and implementation-oriented by introducing measurable quantities such as statistical formulas, ratios, or example computations. This version bridges conceptual factor ideas with practical implementation cues.
\end{itemize}

\subsection{Multi-Agent Quality Checker}
\label{A: Multi-Agent Quality Checker}
As shown in Figure~\ref{fig: CogAlpha}, the Multi-Agent Quality Checker operates through the following sequence:

\paragraph{Code Quality Agent.} It performs the first-pass audit of the raw LLM-generated code. It detects syntactic errors, undefined variables, formatting inconsistencies, invalid library calls, and potential runtime failures using static analysis and lightweight interpreter checks. This agent ensures that the code is structurally well-formed before deeper semantic inspection takes place.

\paragraph{Code Repair Agent.} If the Code Quality Agent identifies issues, the Code Repair Agent attempts to fix them autonomously. Repairs include correcting import statements, rewriting malformed expressions, resolving type mismatches, and rewriting unstable numerical operations. This agent ensures that the factor is at least syntactically and operationally viable.

\paragraph{Judge Agent.}
Once the code is syntactically clean, the Judge Agent evaluates the factor at a semantic level. It assesses whether the factor is:
\begin{itemize}[leftmargin=1em]
    \item \textit{Logically consistent}: correct operator ordering, coherent data flow, no degenerate expressions;
    \item \textit{Technically correct}: valid use of rolling windows, transforms, and TA-Lib functions;
    \item \textit{Economically meaningful}: obeys financial intuition and avoids fabricated indicators.
\end{itemize}
Factors that fail this semantic audit are routed to the Logic Improvement Agent.

\paragraph{Logic Improvement Agent.} This agent refines factors that exhibit weak or inconsistent logic. It restructures formulas, adjusts window parameters, replaces dubious transformations, eliminates redundant operations, and enhances the overall financial interpretability while preserving the original modeling intent. This refinement improves robustness without altering the factor’s core hypothesis.

\paragraph{Execution and Numerical Stability Check.} After passing the logical audits, the code is executed in a restricted sandbox. We evaluate numerical stability by detecting runtime errors, \textit{NaN} propagation, overflow/underflow, invalid logarithms, and unstable normalizations. Codes that fail are rejected or sent back to earlier agents for correction.

\paragraph{Temporal Leakage Unit Test.} Finally, Static Safety conducts a domain-specific leakage test to ensure that the factor does not use future information. This test detects forward-looking shifts (e.g., \texttt{shift(-1)}), misaligned rolling windows, or implicit temporal violations that may pass standard code-safety tools. Only factors with zero leakage are accepted.

\paragraph{Output.} Codes that pass all agents form a pool of safe, executable, and leakage-free alpha factors. These factors constitute the foundation for the next stage, \textit{Thinking Evolution}, which focuses on improving reasoning reliability and logical effectiveness. By enforcing strict correctness at the code level first, CogAlpha ensures that all evolution occurs on top of a solid and trustworthy computational base.

\subsection{Fitness Evaluation}
\label{A: Fitness Evaluation}
The threshold values for the five predictive power metrics may vary depending on the dataset. For example, on the CSI300, we constrain each metric with a minimum bound to prevent the dominance of outliers: 
\textit{IC} and \textit{RankIC} are bounded below by 0.005, 
\textit{ICIR} and \textit{RankICIR} by 0.05, 
and \textit{MI} by 0.02. 
For elite factors, the minimum bounds are slightly higher: 0.01 for \textit{IC} and \textit{RankIC}, 
0.1 for \textit{ICIR} and \textit{RankICIR}, 
and 0.02 for \textit{MI}. 
However, on the S\&P500, we apply similar constraints to prevent the dominance of outliers:
\textit{IC} and \textit{RankIC} are bounded below by 0.005, 
\textit{ICIR} and \textit{RankICIR} by 0.05, 
and \textit{MI} by 0.012. 
For elite factors, the minimum bounds are slightly higher: 0.01 for \textit{IC} and \textit{RankIC}, 
0.1 for \textit{ICIR} and \textit{RankICIR}, 
and 0.012 for \textit{MI}. 
This is because it is harder to mine alpha signals in a more effective stock market. 

Non-linear relationships (e.g., MI) suggest that the market may not be fully efficient, and certain information may not be fully reflected in prices, thus providing opportunities for factor investing. Non-linear factor models are better able to capture complex patterns in the market and may offer opportunities for excess returns, especially when the market is not fully efficient.

\section{Experiments}
\subsection{Datasets}
\label{A: Datasets}
Our experiments are mainly conducted on the CSI300 (China Securities Index 300). Its components consist of 300 large-cap A-share stocks in the Chinese market. We primarily use the 10-day return as the prediction target, with buying and selling at the open price. The dataset is split chronologically into training (2011/01/01-2019/12/31), validation (2020/01/01-2020/12/31), and test (2021/01/01-2024/12/01) periods. On the same dataset, we also use a 30-day return as the prediction target.

CSI500 (China Securities Index 500) consists of 500 relatively smaller but liquid companies in China. We primarily use the 10-day return as the prediction target, with buying and selling at the open price. The dataset is split chronologically into training (2011/01/01-2019/12/31), validation (2020/01/01-2020/12/31), and test (2021/01/01-2024/12/01) periods.

Three other datasets from two different stock markets (U.S. and HK) are also used. The S\&P500 is the Standard \& Poor's 500 Index, and its components include 500 of the largest publicly traded companies in the U.S. stock market. The dataset is split chronologically into training (2007/01/01-2014/12/31), validation (2015/01/01-2015/12/31), and test (2016/01/01-2020/12/01) periods.

The HSI (Hang Seng Index) tracks the performance of the largest and most liquid companies listed on the Hong Kong Stock Exchange. Currently, according to~\cite{AASTOCKS_HKIndex}, it has 89 stocks. The dataset is split chronologically into training (2011/01/01-2019/12/31), validation (2020/01/01-2020/12/31), and test (2021/01/01-2025/12/01) periods.

The HSCI (Hang Seng China Enterprises Index) covers about the top 95th percentile of the total market capitalization of companies listed on the Main Board of the Stock Exchange of Hong Kong. Currently, according to~\cite{HSI_HSCI}, it includes 509 stocks. The dataset is split chronologically into training (2011/01/01-2019/12/31), validation (2020/01/01-2020/12/31), and test (2021/01/01-2025/12/01) periods.

The CSI300, CSI500, and S\&P500 datasets are downloaded from the Qlib platform~\cite{yang2020qlib}. The HSI and HSCI datasets are downloaded from Yahoo Finance~\cite{Aroussi2024}. All backtesting simulations are implemented and executed entirely within the Qlib framework.

\subsection{Backtest}
\label{A: Backtest}
The top-$50$/drop-$5$ strategy is a ranking-based portfolio construction method that selects the top $50$ stocks with the highest predicted returns while limiting daily portfolio turnover. On each trading day, the portfolio retains previously selected high-ranking stocks and replaces at most $5$ positions. All trades are executed at the opening price. The open cost is set to 0.05\%, and the close cost is set to 0.15\%. A minimum transaction fee of 5 CNY is applied to each trade.

\begin{table*}[t]
\centering
\tiny  
\resizebox{\textwidth}{!}{
\begin{tabular}{lcccccccc}
\toprule
\multicolumn{1}{c}{\multirow{2}{*}{\textbf{Models}}}  & \multicolumn{1}{c}{\multirow{2}{*}{\textbf{Dataset}}} & \multicolumn{1}{c}{\multirow{2}{*}{\textbf{Horizon}}} &
\multicolumn{6}{c}{\textbf{CSI300}} \\ 
\cmidrule(lr){4-9}
\multicolumn{1}{c}{} & & & \textbf{IC} & \textbf{RankIC} & \textbf{ICIR} & \textbf{RankICIR} & \textbf{AER} & \textbf{IR} \\ 
\midrule
Agent & \multicolumn{1}{c}{\multirow{5}{*}{CSI300}} & \multicolumn{1}{c}{\multirow{5}{*}{10 days}} & 0.0300 & 0.0318 & 0.2501 & 0.2595 & 0.0789 & 0.8015 \\
Agent\_E & & & 0.0219 & 0.0420 & 0.1932 & 0.3322 & 0.0808 & 0.8999 \\ 
Agent\_EA & & & 0.0315 & 0.0491 & 0.2568 & 0.3583 & 0.0825 & 1.0145 \\ 
Agent\_EAG & & & 0.0414 & 0.0501 & 0.3239 & 0.3599 & 0.1245 & 1.4668 \\ 
Agent\_EAGH (CogAlpha) & & & \textbf{0.0591} & \textbf{0.0814} & \textbf{0.3410} & \textbf{0.4350} & \textbf{0.1639} & \textbf{1.8999} \\ 
\bottomrule
\end{tabular}
}
\caption{
Ablation study of \textsc{CogAlpha}. 
\textbf{A} denotes Adaptive Generation, 
\textbf{G} denotes Diversified Guidance, 
\textbf{H} denotes the Seven-Level Agent Hierarchy, 
and \textbf{E} denotes Thinking Evolution.
The best performance values for each task are highlighted in \textbf{bold}.}
\label{T: Ablation}
\vspace{-0.4cm}
\end{table*}

\subsection{Metrics}
\label{A: Metrics}
We use five factor predictive power metrics: Information Coefficient (\textbf{IC}), Information Coefficient Information Ratio (\textbf{ICIR}),  Rank Information Coefficient (\textbf{RankIC}), Rank Information Coefficient Information Ratio (\textbf{RankICIR}), and Mutual Information (\textbf{MI}). Assume there are \(N_t\) assets at time \(t\). Let \(f_{i,t}\) represent the predicted returns for asset \(i\) at time \(t\), and \(r_{t+1}\) represent the total return over the subsequent period, from \(t\) to \(t + 1\). The evaluation spans over \(T\) time periods. We also use two performance metrics: AER and IR.

The \textbf{Information Coefficient (IC)} measures the linear correlation between factor values and subsequent total returns. It is the average of each linear cross-sectional
relationship between factor values and subsequent returns at time \(t\) over all \(T\) periods:
\begin{equation}
\footnotesize
\begin{aligned}
\mathrm{IC} &= \frac{1}{T} \sum_{t=1}^{T} \mathrm{IC}_t \\
\mathrm{IC}_t &=
\frac{\sum_{i=1}^{N_t} (f_{i,t} - \bar{f}_t)(r_{i,t+1} - \bar{r}_{t+1})
}{
\sqrt{\sum_{i=1}^{N_t} (f_{i,t} - \bar{f}_t)^2}
\sqrt{\sum_{i=1}^{N_t} (r_{i,t+1} - \bar{r}_{t+1})^2}
}
\label{eq:ic}
\end{aligned}
\end{equation}

The \textbf{Information Coefficient Information Ratio (ICIR)} evaluates the stability of \textbf{IC} across time:
\begin{equation}
\mathrm{ICIR}
=
\frac{\mathbb{E}[\mathrm{IC}_t]}{\mathrm{Std}[\mathrm{IC}_t]}
\;\approx\;
\frac{\mathrm{IC}}{\mathrm{Std}(\{\mathrm{IC}_t\}_{t=1}^{T})},
\end{equation}
where $\mathrm{IC}$ denotes the time-averaged $\mathrm{IC}_t$.

The \textbf{Rank Information Coefficient (RankIC)} measures the monotonic
relationship between the factor and the subsequent total returns. Let
\[
u_{i,t} = \mathrm{rank}(f_{i,t}), \qquad
v_{i,t} = \mathrm{rank}(r_{i,t+1}),
\]
and their means $\bar{u}_t,\,\bar{v}_t$ across $N_t$ assets. Analogous to IC, RankIC over period \(T\) can be expressed as:
\begin{equation}
\footnotesize
\begin{aligned}
\mathrm{RankIC} &= \frac{1}{T} \sum_{t=1}^{T} \mathrm{RankIC}_t \\
\mathrm{RankIC}_t &= \frac{
\sum_{i=1}^{N_t}(u_{i,t}-\bar{u}_t)(v_{i,t}-\bar{v}_t)
}{
\sqrt{\sum_{i=1}^{N_t}(u_{i,t}-\bar{u}_t)^2}
\sqrt{\sum_{i=1}^{N_t}(v_{i,t}-\bar{v}_t)^2}
}.
\label{eq:rankic}
\end{aligned}
\end{equation}

Analogous to ICIR, the \textbf{RankIC Information Ratio (RankICIR)} measures the temporal stability of
RankIC:
\begin{equation} \footnotesize
\mathrm{RankICIR}
=
\frac{\mathbb{E}[\mathrm{RankIC}_t]}{\mathrm{Std}[\mathrm{RankIC}_t]}
\;\approx\;
\frac{\overline{\mathrm{RankIC}}}{\mathrm{Std}(\{\mathrm{RankIC}_t\}_{t=1}^{T})}.
\end{equation}

The \textbf{Mutual Information (MI)} captures the nonlinear dependence between
factor values and the subsequent total returns. It measures the reduction in uncertainty of
$R$ given knowledge of $F$:
\begin{equation} \footnotesize
\mathrm{MI}(F, R)
=
\iint p(f, r)\,
\log\frac{p(f, r)}{p(f)\,p(r)}\,df\,dr,
\end{equation}
where $p(f, r)$ denotes the joint density of factor $f$ and return $r$, and
$p(f)$, $p(r)$ are their respective marginal densities. A higher MI implies a
stronger (possibly nonlinear) dependency between the factor and subsequent
returns.

\paragraph{Annualized Excess Return (AER).}
Following Qlib's implementation, we compute daily excess returns as
\[
r_t = r^{\text{port}}_t - r^{\text{bench}}_t - \text{cost}_t,
\]
where $r^{\text{port}}_t$ is the portfolio return, $r^{\text{bench}}_t$ is the
benchmark return, and $\text{cost}_t$ is the transaction cost. The average
daily excess return is
\[
\mu = \frac{1}{T} \sum_{t=1}^{T} r_t,
\]
and the annualized excess return is obtained via arithmetic scaling:
\[
\mathrm{AER} = \mu \times N,
\]
where $N$ is the number of trading periods in a year (e.g., $N = 252$ for daily returns).

\paragraph{Information Ratio (IR).}
The standard deviation of daily excess returns is
\[
\sigma = \sqrt{\frac{1}{T-1} \sum_{t=1}^{T} (r_t - \mu)^2 }.
\]
Qlib annualizes the Information Ratio using
\[
\mathrm{IR} = \frac{\mu}{\sigma}\sqrt{N}.
\]

\subsection{Training Setting}
The size of the initial pool is set to 80, meaning that the minimum number of alphas generated by the task-specific agents is 80. 
The parent pool size is set to 32, indicating that after filtering, at most 32 alphas are retained and passed to the next generation. 
The children's pool is set to be three times the size of the parent pool, meaning that the minimum number of alphas generated by the evolutionary agents is 96. 
By default, each task-specific agent leads a complete evolutionary cycle, which consists of 24 generations and 3 inner sub-cycles, with each sub-cycle comprising 8 generations. 
Thus, each task-specific agent initiates the evolutionary search 3 times.  
In addition, every 2 generations, new alphas generated by the task-specific agents are filtered and injected into the parent pool. 
For each generation, the top two elite alphas from the previous generation are always carried forward to the next. 
All alphas containing more than 30\% NaN values or failing the multi-agent quality checker are discarded. 
All experiments are conducted on NVIDIA H100 GPUs.

We use rolling training with a rolling step of 126. Two models are employed in this work: the \textit{LGBMRegressor} and \textit{Ridge} models. The \textit{LGBMRegressor} is configured with a learning rate of 0.0001, 32 leaves per tree, a maximum depth of 12, and regularization terms (\texttt{reg\_alpha} and \texttt{reg\_lambda}) set to 1.0. The model uses a total of 1000 trees with sampling techniques (feature and bagging fractions set to 0.8) to reduce overfitting. For the \textit{Ridge} model, the regularization strength (\texttt{alpha}) is set to 10, which controls the regularization applied to the model to prevent overfitting. 

We employ two stopping conditions for the evolutionary process. By default, evolution runs for a fixed number of predefined steps. We additionally incorporate a plateau-based early stopping rule, where we track the elite-pool performance and compute the improvement between two consecutive windows of length $\textit{plateau\_win}$, defined as
\[
\delta = \mathrm{mean}(\text{curr}) - \mathrm{mean}(\text{prev}).
\]
If $\delta \leq 0.001$, the evolution for that island or run is terminated.

\subsection{Computational Cost}
On the CSI300 dataset, generating a single alpha factor typically takes 5--9 seconds, and completing one generation takes approximately 1 hour. 

For comparison, deep learning models running on GPU exhibit the following training times: CNN, GRU, and LSTM models require around 20 minutes, while Transformer-based models take approximately 40 minutes. For traditional machine learning models running on CPU, AdaBoost requires around 6 hours, Random Forest takes approximately 40 minutes, LightGBM completes in about 2 minutes, and linear models typically require only 5--10 seconds. 

For all main experiments, we use a single H100 GPU. The evolutionary process is conducted using a local model (\texttt{gpt-oss-120b}), which incurs no API cost.

\subsection{Randomness of LLMs}
Due to the inherent randomness in the outputs of large models, the results may vary with each run. However, factor mining is different from other experiments in that good factors can be accumulated and stored. Therefore, the experimental results presented in this paper reflect the outcomes after a single round of factor mining. 
% We evaluate the mined alphas using three independent runs with different random seeds and report the median value across runs.

\subsection{Ablation Study}
\label{A: Ablation Study}
We evaluate the effectiveness of each component of CogAlapha: Adaptive Generation (\textbf{A}), Diversified Guidance (\textbf{G}), Seven-Level Agent Hierarchy (\textbf{H}), and Thinking Evolution (\textbf{E}). As shown in Table~\ref{T: Ablation}, the four parts can, to some extent, improve the effectiveness and performance of alpha mining.

\subsection{Hyperparameter Design and Analysis}

Our hyperparameter design is inspired by two prior works~\cite{lee2025evolving, zhang2025verbalized}. In~\cite{lee2025evolving}, the parent pool size is set to 5 for each island/agent, with 10 generations per cycle. In~\cite{zhang2025verbalized}, it is suggested that allowing LLMs to generate a batch of responses (e.g., 5) at once improves diversity.

In our setting, we employ 21 heterogeneous agents and adopt the golden ratio, which is commonly used in quantitative finance for balanced allocation, to randomly select 13 agents for constructing the initial factor pool. Following~\cite{lee2025evolving, zhang2025verbalized}, each selected agent generates approximately 5--6 alpha factors, resulting in an initial pool of around 80 factors. We then apply the golden ratio again to form a parent pool of size 32. Given three distinct evolution operators, the resulting children pool size is $3 \times 32 = 96$.

To analyze the impact of hyperparameters, we conduct ablation experiments on the \texttt{AgentMarketCycle} agent under different configurations. We denote each configuration as \texttt{P$\cdot$\_G$\cdot$\_H$\cdot$}, where \texttt{P} is the parent pool size, \texttt{G} is the number of generations per cycle, and \texttt{H} is the length of sub-cycles. The results on the CSI300 dataset with a 10-day horizon are summarized in Table~\ref{tab:hyperparam}.

\begin{table}[t]
\centering
\small
\begin{tabular}{lcccc}
\toprule
Configuration & IC & RankIC & ICIR & RankICIR \\
\midrule
P16\_G24\_H8  & 0.0362 & 0.0508 & \textbf{0.3168} & 0.3805 \\
P32\_G24\_H8  & 0.0315 & 0.0475 & 0.2364 & 0.3379 \\
P48\_G24\_H8  & 0.0199 & 0.0285 & 0.1835 & 0.2480 \\
P32\_G24\_H2  & \textbf{0.0394} & \textbf{0.0625} & 0.3128 & \textbf{0.4364} \\
P32\_G24\_H4  & 0.0281 & 0.0477 & 0.2309 & 0.3507 \\
P32\_G24\_H12 & 0.0340 & 0.0524 & 0.2928 & 0.3759 \\
P32\_G8\_H8   & 0.0283 & 0.0447 & 0.2413 & 0.3542 \\
P32\_G16\_H8  & 0.0326 & 0.0433 & 0.2505 & 0.3458 \\
\bottomrule
\end{tabular}
\caption{Hyperparameter analysis on CSI300 (10-day horizon).}
\label{tab:hyperparam}
\end{table}

From the results, the configuration \texttt{P32\_G24\_H2} achieves the best overall performance. Nevertheless, other configurations may still yield competitive results and are capable of discovering effective alpha factors. We therefore argue that there is no universally optimal hyperparameter setting across different time periods and market conditions. Instead, maintaining diversity in hyperparameter configurations is beneficial for more comprehensive alpha factor discovery.

\subsection{Evolution of Alphas}
\label{A: Evolution of Alphas}
To demonstrate the evolution capability of \textbf{CogAlpha}, we show an example of how liquidity-related alphas evolve over multiple iterations. Each generated alpha is evaluated by predictive metrics (IC and RankIC). Poorly performing alphas are automatically filtered out, while stronger ones are preserved and further evolved.

\begin{lstlisting}[caption={Mutated alpha variant using full price range}, label={lst: alpha_code2}, basicstyle=\tiny]
def factor_dayhigh_impact_per_vol(df):
    """Price-impact proxy: (high-low) per unit of volume.
    Larger values indicate that price moves a lot while little volume trades, signalling thin liquidity."""
    df_copy = df.copy()
    df_copy['price_range'] = df_copy['high'] - df_copy['low']
    df_copy['factor_dayhigh_impact_per_vol'] = df_copy['price_range'] / (df_copy['volume'] + 1e-9)
    return df_copy['factor_dayhigh_impact_per_vol']
\end{lstlisting}

The first version (\autoref{lst: alpha_code1}) represents an initial manually designed alpha. It measures the liquidity impact as the price rise \((high - close)\) per unit of traded volume. Its metrics are: \textbf{IC: 0.0090}, \textbf{RankIC: 0.0061}. Through mutation, the model generates an alternative formulation (\autoref{lst: alpha_code2}) that uses the full daily price range \((high - low)\) instead of the closing difference. This captures broader intraday liquidity behavior. Its metrics slightly decrease to \textbf{IC: 0.0073}, \textbf{RankIC: 0.0021}, and thus this version is discarded in later rounds. 

\begin{figure*}[t] 
    \centering
    \footnotesize
    % \fbox{\rule[-.5cm]{0cm}{4cm} \rule[-.5cm]{4cm}{0cm}}
    \includegraphics[width=\linewidth]{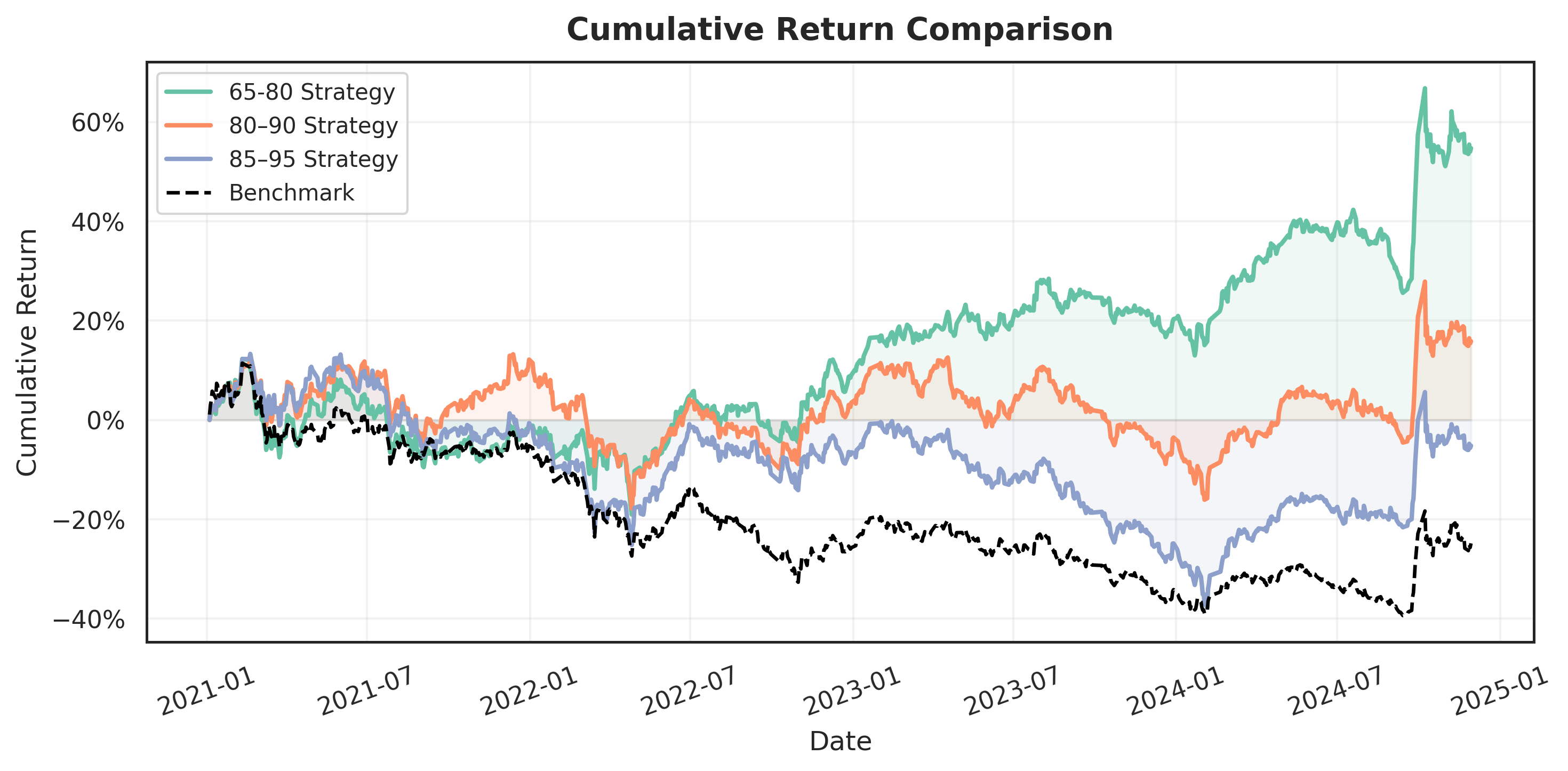}
    % \vspace{-0.4cm}
    % \fbox{\rule{0pt}{2.5in} \rule{0.9\linewidth}{0pt}}
    \caption{\textbf{Performance of CogAlpha on different fitness threshold} 
    }
    \label{fig: Cumulative Return Different Threshold}
    % \vspace{-0.3cm}
\end{figure*}

\begin{lstlisting}[caption={Evolved alpha after multi-round optimization}, label={lst: alpha_code3}, basicstyle=\tiny]
def factor_price_impact_per_vol_tanh_1d(df):
    """Impact proxy: absolute daily price move per dollar volume.
    Steps:
    1) Compute absolute price move (|Close-Open|).
    2) Compute dollar volume (Volume*Close).
    3) Form raw impact = absolute move / (dollar volume + ε).
    4) Apply tanh to bound the factor within (-1, 1)."""
    df_copy = df.copy()
    eps = 1e-9
    df_copy.loc[:, "abs_move"] = (df_copy["close"] - df_copy["open"]).abs()
    df_copy.loc[:, "dollar_vol"] = df_copy["volume"] * df_copy["close"]
    df_copy.loc[:, "raw_impact"] = df_copy["abs_move"] / (df_copy["dollar_vol"] + eps)
    df_copy.loc[:, "factor_price_impact_per_vol_tanh_1d"] = np.tanh(df_copy["raw_impact"])
    return df_copy["factor_price_impact_per_vol_tanh_1d"]
\end{lstlisting}

After several evolutionary rounds, CogAlpha produces a more refined version (\autoref{lst: alpha_code3}). It normalizes the absolute daily price move by dollar volume and applies a $\tanh$ transformation to ensure boundedness and robustness. The evolved alpha achieves significantly improved performance with \textbf{IC: 0.0141} and \textbf{RankIC: 0.0087}, showing the ability of the evolutionary mechanism to refine quantitative factors effectively.

\begin{table*}[t]
\centering
\tiny
\resizebox{\textwidth}{!}{
\begin{tabular}{ccccccccc}
\toprule
\textbf{Dataset} & \textbf{Horizon} & \textbf{Models} & \textbf{Training Method} &
\textbf{IC} & \textbf{RankIC} & \textbf{ICIR} & \textbf{RankICIR} \\
\midrule

\multirow{10}{*}{CSI300}
 & \multirow{5}{*}{10 days} 
 & CNN & CNN & 0.0268 & 0.0392 & 0.2432 & 0.3117 \\
 & & Linear & Linear & 0.0165 & 0.0211 & 0.1612 & 0.1655 \\
 & & XGBoost & XGBoost & 0.0257 & 0.0376 & 0.2783 & 0.4093 \\
 & & CogAlpha & Ridge & 0.0539 & 0.0714 & \textbf{0.3471} & \textbf{0.4380} \\
 & & CogAlpha & LightGBM & \textbf{0.0591} & \textbf{0.0814} & 0.3410 & 0.4350 \\
 \cmidrule(lr){2-8}
 & \multirow{5}{*}{30 days} 
 & CNN & CNN & 0.0445 & 0.0644 & 0.4118 & 0.6023 \\
 & & Linear & Linear & 0.0222 & 0.0482 & 0.2174 & 0.4218 \\
 & & XGBoost & XGBoost & 0.0402 & 0.0507 & 0.4181 & 0.5896 \\
 & & CogAlpha & Ridge & 0.0815 & 0.1069 & 0.4408 & 0.5403 \\
 & & CogAlpha & LightGBM & \textbf{0.0886} & \textbf{0.1243} & \textbf{0.4933} & \textbf{0.6740} \\

\midrule
\multirow{4}{*}{CSI500}
 & \multirow{4}{*}{10 days} 
 & CNN & CNN & 0.0353 & 0.0490 & 0.2615 & 0.3959 \\
 & & Linear & Linear & 0.0273 & 0.0391 & 0.2787 & 0.3671 \\
 & & XGBoost & XGBoost & 0.0282 & 0.0290 & 0.2458 & 0.3046 \\
 & & CogAlpha & Ridge & \textbf{0.0455} & \textbf{0.0738} & \textbf{0.2903} & \textbf{0.4752} \\

\midrule
\multirow{4}{*}{S\&P500}
 & \multirow{4}{*}{10 days} 
 & CNN & CNN & -0.0040 & -0.0028 & -0.0508 & -0.0325 \\
 & & Linear & Linear & 0.0031 & 0.0004 & 0.0408 & 0.0045 \\
 & & XGBoost & XGBoost & -0.0020 & 0.0045 & -0.0434 & 0.0670 \\
 & & CogAlpha & Ridge & \textbf{0.0217} & \textbf{0.0226} & \textbf{0.2189} & \textbf{0.1726} \\

\midrule
\multirow{4}{*}{HSI}
 & \multirow{4}{*}{10 days} 
 & CNN & CNN & 0.0218 & 0.0177 & 0.1603 & 0.1248 \\
 & & Linear & Linear & 0.0218 & 0.0158 & 0.1647 & 0.0967 \\
 & & XGBoost & XGBoost & -0.0031 & 0.0003 & -0.0272 & 0.0026 \\
 & & CogAlpha & Ridge & \textbf{0.0327} & \textbf{0.0400} & \textbf{0.1903} & \textbf{0.2330} \\

\midrule
\multirow{4}{*}{HSCI}
 & \multirow{4}{*}{10 days} 
 & CNN & CNN & 0.0385 & 0.0242 & 0.4006 & 0.2304 \\
 & & Linear & Linear & 0.0171 & 0.0142 & 0.2258 & 0.1401 \\
 & & XGBoost & XGBoost & 0.0295 & 0.0144 & 0.3153 & 0.1690 \\
 & & CogAlpha & Ridge & \textbf{0.0562} & \textbf{0.0495} & \textbf{0.5396} & \textbf{0.4394} \\

\bottomrule
\end{tabular}
}
\caption{Performance comparison of \textsc{CogAlpha} across different datasets, horizons, and training methods. Results are grouped by dataset, with IC, RankIC, ICIR, and RankICIR reported for each configuration. The best performance values for each task are highlighted in \textbf{bold}.}
\label{T: Different Settings}
\end{table*}

After a complete evolution cycle, CogAlpha is able to generate a large number of single-factor alphas with strong predictive power, many of which achieve \textbf{absolute IC values above 0.05} and \textbf{absolute RankIC values above 0.07}. This demonstrates the framework's capacity to autonomously explore and optimize factor space toward higher-performing and more interpretable alphas. The following are a few examples of elite alphas:

\paragraph{factor\_lownorm\_slopecos\_30d\_low}:
\textit{Train metrics (raw)}: IC = -0.0498, RankIC = -0.0791, ICIR = -0.3416, RankICIR = -0.5016;
\textit{Test metrics (Ridge)}: IC = 0.0507, RankIC = 0.0704, ICIR = 0.3116, RankICIR = 0.4262
% Train metircs: 'IC': -0.0498, 'RankIC': -0.0791, 'ICIR': -0.3416, 'RankICIR': -0.5016
% Test metrics: 'IC': 0.0507, 'RankIC': 0.0704, 'ICIR': 0.3116, 'RankICIR': 0.4262
\begin{lstlisting}[caption={Evolved alpha after multi-round optimization}, label={lst: alpha_code4}, basicstyle=\tiny]
def factor_lownorm_slopecos_30d_low(df):
    """Low-price relative to its 30-day EMA multiplied by a cycle-aligned slope*cosine.
    Steps:
    1. Normalise the low price by its 30-day EMA -> dynamic support level.
    2. Estimate the soft-sign slope of log-price vs. log-volume (EWMA cov/var) -> short-term trend.
    3. Compute the EMA-12/-48 cosine to capture the dominant market cycle phase.
    4. Combine slope and cosine (slope*cosine) as a single "cycle factor".
    5. Multiply the normalised low by the cycle factor; the product is high when price is near support and the cycle direction is favourable."""
    df_copy = df.copy()
    eps = 1e-9

    # 1. Normalised low price (support level)
    low_ema30 = df_copy['low'].ewm(span=30, adjust=False).mean()
    norm_low = df_copy['low'] / (low_ema30 + eps)

    # 2. Soft-sign slope of log-price vs. log-volume
    log_ret = np.log(df_copy['close'] / df_copy['close'].shift(1))
    log_vol = np.log(df_copy['volume'] / df_copy['volume'].shift(1))
    cov = log_ret.ewm(halflife=20, adjust=False).cov(log_vol)
    var = log_vol.ewm(halflife=20, adjust=False).var() + eps
    slope_raw = cov / var
    slope = slope_raw / (1.0 + slope_raw.abs())

    # 3. EMA-12/-48 cosine (cycle phase)
    ema12 = df_copy['close'].ewm(span=12, adjust=False).mean()
    ema48 = df_copy['close'].ewm(span=48, adjust=False).mean()
    cos_phase = ema12 / np.sqrt(ema12**2 + ema48**2 + eps)

    # 4. Cycle factor = slope * cosine
    cycle_factor = slope * cos_phase

    # 5. Final factor
    df_copy['factor_lownorm_slopecos_30d_low'] = norm_low * cycle_factor

    return df_copy['factor_lownorm_slopecos_30d_low']
\end{lstlisting}

% Train metircs: 'IC': -0.0473, 'RankIC': -0.0668, 'ICIR': -0.3749, 'RankICIR': -0.4473
% Test metrics: 'IC': 0.0491, 'RankIC': 0.069, 'ICIR': 0.2717, 'RankICIR': 0.3604
\paragraph{factor\_pressure\_drawdown\_fisher\_10d}:
\textit{Train metrics (raw)}: IC = -0.0473, RankIC = -0.0668, ICIR = -0.3749, RankICIR = -0.4473;
\textit{Test metrics (LightGBM)}: IC = 0.0491, RankIC = 0.069, ICIR = 0.2717, RankICIR = 0.3604

\begin{lstlisting}[caption={Evolved alpha after multi-round optimization}, label={lst: alpha_code5}, basicstyle=\tiny]
def factor_pressure_drawdown_fisher_10d(df):
    """Pressure median * Fisher-corr(20d) * drawdown EMA-8.
    1. pressure = |C-O|*log1p(V), median over 10 days.
    2. drawdown = - (C - rolling_max_252) / rolling_max_252, EMA-8 smoothed.
    3. fisher = Fisher-transform of 20-day corr(log returns, log volume).
    4. factor = pressure_med10 * fisher_corr * tanh(decayed_dd)."""
    df_copy = df.copy()
    eps = 1e-12

    # 1. pressure median (10 d)
    pressure_raw = (df_copy['close'] - df_copy['open']).abs() * np.log1p(df_copy['volume'])
    pressure_med10 = pressure_raw.rolling(window=10, min_periods=10).median()

    # 2. EMA-8 drawdown magnitude
    roll_max = df_copy['close'].rolling(window=252, min_periods=1).max()
    drawdown = -((df_copy['close'] - roll_max) / (roll_max + eps))
    decayed_dd = drawdown.ewm(span=8, adjust=False).mean()

    # 3. Fisher-transformed 20-day return-volume correlation
    log_ret = np.log(df_copy['close'] + eps).diff()
    log_vol = np.log(df_copy['volume'] + eps).diff()
    corr20 = log_ret.rolling(window=20, min_periods=20).corr(log_vol)
    fisher_corr = np.arctanh(corr20.clip(-0.999, 0.999))

    # 4. Combine
    factor = pressure_med10 * fisher_corr * np.tanh(decayed_dd)

    factor_name = 'factor_pressure_drawdown_fisher_10d'
    factor.name = factor_name
    df_copy[factor_name] = factor
    return df_copy[factor_name]
\end{lstlisting}

% Train metircs: 'IC': -0.0552, 'RankIC': -0.0742, 'ICIR': -0.475, 'RankICIR': -0.5141
% Test metrics: 'IC': 0.0503, 'RankIC': 0.0663, 'ICIR': 0.3017, 'RankICIR': 0.392
\paragraph{factor\_herd\_drawdown\_synergy\_gate\_ema10}:
\textit{Train metrics (raw)}: IC = -0.0552, RankIC = -0.0742, ICIR = -0.475, RankICIR = -0.5141;
\textit{Test metrics (LightGBM)}: IC = 0.0503, RankIC = 0.0663, ICIR = 0.3017, RankICIR = 0.392
   
\begin{lstlisting}[caption={Evolved alpha after multi-round optimization}, label={lst: alpha_code6}, basicstyle=\tiny]
def factor_herd_drawdown_synergy_gate_ema10(df):
    """Herd-drawdown synergy with a price-trend gate and a stability filter.
    Steps (5):
    1) Core herd signal = tanh(herd_corr_retvol_20d * herd_fisher_20d_rollingwinsor).  
       Tanh bounds the product, reducing extreme values.
    2) Trend gate = sign(C-O) * tanh((C-O) / sigma20) where sigma20 is the 20-day rolling std of close.
       Captures directionality while limiting influence of noisy price moves.
    3) Raw synergy = core * drawdown_steadiness_energy_tanh_regime_10d * trend_gate.  
       Combines herd, drawdown energy, and price direction.
    4) Stability = 1 / (1 + MAD20(body_ratio)) with body_ratio = (C-O)/(H-L+ε).  
       MAD provides a robust dispersion measure; the stability term down-weights volatile candles.
    5) EMA-10 (ticker-wise) smooths raw_synergy * stability, yielding a stable, lagged factor."""
    df_copy = df.copy()
    eps = np.finfo(float).eps

    # 1. bounded herd core
    core = np.tanh(df_copy['factor_herd_corr_retvol_20d'] *
                   df_copy['factor_herd_fisher_20d_rollingwinsor'])

    # 2. asymmetric price-trend gate (20-day close std per ticker)
    price_change = df_copy['close'] - df_copy['open']
    std20 = df_copy['close'].groupby(level='ticker').transform(
        lambda s: s.rolling(window=20, min_periods=1).std()
    )
    trend_gate = np.sign(price_change) * np.tanh(price_change / (std20 + eps))

    # 3. raw synergy with drawdown-energy factor
    raw_synergy = core * df_copy['factor_drawdown_steadiness_energy_tanh_regime_10d'] * trend_gate

    # 4. stability gate via MAD of body ratio
    body_ratio = price_change / (df_copy['high'] - df_copy['low'] + eps)
    mad20 = body_ratio.groupby(level='ticker').transform(
        lambda s: s.rolling(window=20, min_periods=10).apply(
            lambda w: np.median(np.abs(w - np.median(w))), raw=False)
    )
    stability = 1.0 / (1.0 + mad20)

    # 5. EMA-10 smoothing per ticker, preserving (date, ticker) index
    synergy = raw_synergy * stability
    result = synergy.groupby(level='ticker', group_keys=False).apply(
        lambda s: s.ewm(span=10, adjust=False).mean()
    )
    result.name = 'factor_herd_drawdown_synergy_gate_ema10'

    df_copy['factor_herd_drawdown_synergy_gate_ema10'] = result
    return df_copy['factor_herd_drawdown_synergy_gate_ema10']
\end{lstlisting}

\subsection{Different Fitness Threshold}
\label{A: Different Fitness Threshold}

We analyze the sensitivity of our method to different threshold settings used for filtering alpha factors. 
To maintain the quality of the selected alphas, we experiment with three threshold pairs: (65, 80), (80, 90), and (85, 95). 
In each pair, the former value represents the percentile threshold for \textit{qualified factors} that advance to the next generation, while the latter corresponds to the percentile threshold for \textit{elite factors} that are directly stored in the final candidate pool. 
For comparison, we also establish a baseline configuration to ensure the quality consistency of filtered alphas. 
Specifically, thresholds for each predictive metric are determined based on the empirical distribution of factor scores. 
We constrain each metric by a minimum bound to prevent the dominance of outliers:
\textit{IC} and \textit{RankIC} are bounded below by 0.005, 
\textit{ICIR} and \textit{RankICIR} by 0.05, 
and \textit{MI} by 0.02. 
For elite factors, the minimum bounds are slightly higher: 0.01 for \textit{IC} and \textit{RankIC}, 
0.1 for \textit{ICIR} and \textit{RankICIR}, 
and 0.02 for \textit{MI}. 
As shown in Figure~\ref{fig: Cumulative Return Different Threshold}, the threshold pair (65, 80) yields better overall performance. 
This result may be attributed to the larger parent pool size under this configuration, which encourages evolutionary search to explore a broader alpha space and mitigates the risk of premature convergence to local optima.

\subsection{Generalization to Different Settings}
\label{A: Different Settings}
We test the generalization of CogAlpha on different datasets (CSI300, CSI500, S\&P500, HSI, HSCI), training methods (LightGBM, Ridge), and horizons (10 days, 30 days). As shown in Table~\ref{T: Different Settings}, our method consistently performs well across different settings.

\begin{itemize}
    \item The CSI300 dataset is split chronologically into training (2011/01/01-2019/12/31), validation (2020/01/01-2020/12/31), and test (2021/01/01-2024/12/01) periods.
    \item The CSI500 dataset is split chronologically into training (2011/01/01-2019/12/31), validation (2020/01/01-2020/12/31), and test (2021/01/01-2024/12/01) periods.
    \item The S\&P500 dataset is split chronologically into training (2007/01/01-2014/12/31), validation (2015/01/01-2015/12/31), and test (2016/01/01-2020/12/01) periods. 
    \item The HSI dataset is split chronologically into training (2011/01/01-2019/12/31), validation (2020/01/01-2020/12/31), and test (2021/01/01-2025/12/01) periods.
    \item The HSCI dataset is split chronologically into training (2011/01/01-2019/12/31), validation (2020/01/01-2020/12/31), and test (2021/01/01-2025/12/01) periods.
\end{itemize}

\newpage

\section{Prompt Design}

\subsection{Seven-Level Agent Hierarchy}
\begin{tcolorbox}[
  colback=gray!5, 
  colframe=gray!80,
  title style={fontupper=\bfseries\large},
  title=Seven-Level Agent Hierarchy – Base Agent,
  breakable
]
You are a senior quantitative factor engineer. Below is the schema of the input DataFrame and a list of \textbf{\{columns\_num\}} existing factors: \\

\textbf{\{columns\_desc\}} \\

The input DataFrame consists of \textbf{daily aggregated factors} — i.e., each row represents a single trading day's features for a given stock, already aggregated to daily frequency.

Please generate \textbf{\{num\_per\_request\}} new and original quantitative factor functions that are distinct from the existing ones. Each factor should be implemented as a complete Python function.

\vspace{1mm}
---
\vspace{1mm}

\textbf{\#\#\# Analysis of Effective Factors and Innovation Directions:} \\
Below is a condensed CoT-style summary built from recent successful cases, explaining why they work well. \\
Mini-Chain from Survivors (Observation → Cause → Fix): \\
\textbf{\{effective\_CoT\}} \\

Based on these strengths, focus on incorporating similar principles in new factor creation. 
Seek innovative methods to generate more efficient, robust, and adaptable factors, ensuring they work well in diverse market conditions while avoiding look-ahead/leakage and redundancy.

\vspace{1mm}
---
\vspace{1mm}

\textbf{\#\#\# Analysis of Ineffective Factors and Innovation Directions:} \\
Below is a condensed CoT-style summary built from recent failure cases, explaining why they fail. \\
Mini-Chain from Failures (Observation → Cause → Fix): \\
\textbf{\{ineffective\_CoT\}} \\

Based on these failures, focus on avoiding similar issues in new factor creation. 
Seek innovative methods to generate more effective, robust, and adaptable factors, ensuring they work well in diverse market conditions.

\vspace{1mm}
---
\vspace{1mm}

\textbf{\#\#\# Requirements:} \\
- The input `DataFrame` has a MultiIndex of (date, ticker), and has already been grouped by ticker:
\begin{itemize}
    \item Each input `DataFrame` is a time series of a single stock.
\end{itemize} 

- Output: A `pd.Series` indexed by `(date, ticker)` with the \textbf{same name} as the function. \\

- Each function must:
    \begin{itemize}
        \item Have a descriptive, unique name: \\
            \textit{factor\_\textless logic\textgreater\_\textless transformation(s)\textgreater\\
            \_\textless window(s)\textgreater\_\textless field\textgreater}.
        \item Include a clear docstring explaining the logic and formula.
        \item Balance predictive power with economic/financial interpretability. 
        \item Use an output column name that exactly matches the function name. 
        \item Be concise, precise, and readable. 
        \item Build new alpha factors based on existing ones.
    \end{itemize}

\vspace{1mm}
---
\vspace{1mm}

\textbf{\#\#\# Factor Design Guidance:} \\
You are encouraged to explore a wide variety of signals and techniques related to \{factor\_type\}, including but not limited to: 
\begin{itemize}
    \item List of common techniques / example categories
    \item List of possible interactions or advanced ideas
\end{itemize}

Please do NOT limit yourself to simple formulas or common patterns.  
You are expected to innovate, introduce mathematically sophisticated or unconventional structures, and combine multiple concepts where reasonable.

\vspace{1mm}
The goal is to generate factors that are \textbf{predictive}, \textbf{robust}, and \textbf{economically interpretable}, while being \textbf{structurally diverse} from existing factors.

\vspace{1mm}
---
\vspace{1mm}

\textbf{\#\#\# Pre-imported libraries you can use (current versions):}
\begin{itemize}
\item \texttt{"np"}: \texttt{import numpy as np} \ (numpy version: 2.2.6)
\item \texttt{"pd"}: \texttt{import pandas as pd} \ (pandas version: 2.2.3)
\item \texttt{"stats"}: \texttt{from scipy import stats} \ (scipy version: 1.15.3)
\item \texttt{"talib"}: \texttt{import talib} \ (talib version: 0.5.1)
\item \texttt{"math"}: \texttt{import math} \ (built-in module)
\end{itemize}

\bigskip
\noindent\textbf{Coding Guidelines:}
\begin{itemize}
    \item Ensure the code is robust, efficient, and optimized:
    \begin{itemize}
        \item Handle edge cases and exceptions (e.g., NaN values). 
        \item Minimize unnecessary computations and prefer vectorized operations (e.g., pandas, numpy). 
        \item Ensure numerical stability. 
        \item \textbf{Strict Rule: Nested loops are absolutely forbidden.} 
        \begin{itemize}
            \item You must \textbf{never} write any form of loop inside another loop. 
            \item Forbidden patterns include but are not limited to: 
            \begin{itemize}
                \item \texttt{for} inside \texttt{for}
                \item \texttt{while} inside \texttt{while}
                \item \texttt{for} inside \texttt{while}
                \item \texttt{while} inside \texttt{for}
            \end{itemize}
            \item Any nested iteration structure is \textbf{prohibited}, regardless of indentation depth. 
            \item The use of \texttt{while True} or any potentially infinite loop is \textbf{strictly prohibited}. 
        \end{itemize}
    \end{itemize}
    \item When filtering or assigning values in a DataFrame, always use \texttt{df\_copy.loc[row\_indexer, col\_indexer] = value}.
    \item Code should be clean, maintainable, and efficient for large datasets:
    \begin{itemize}
        \item Use descriptive variable names and minimize memory usage.
        \item Avoid creating unnecessary copies of large DataFrames.
    \end{itemize}
\end{itemize}

\vspace{1mm}
---
\vspace{1mm}

\textbf{\#\#\# Output format specification:} 
\begin{itemize}
    \item Do NOT use markdown (like \texttt{```python}).
    \item Do NOT add any explanation or comments outside the function.
    \item Each function must be wrapped inside: \texttt{\textless\textless function N\textgreater\textgreater} ... \texttt{\textless</function N\textgreater>}.
    \item All generated code must be executable and numerically stable.
    \item Always define intermediate columns (e.g., \texttt{df\_copy['x']}) before referencing them later.
    \item The returned Series \textbf{must} be named exactly the same as the function name.
    \item Each function should follow this format:
    \begin{lstlisting}[language=python]
<<function N>>
def factor_xyz(df):
    """Explain the logic. One clear idea. Short formula. No redundant stacking."""
    df_copy = df.copy()
    # factor computation
    return df_copy["factor_xyz"]
<</function N>>
    \end{lstlisting}
\end{itemize}
\end{tcolorbox}

\begin{tcolorbox}[
  colback=gray!5, 
  colframe=gray!80,
  title style={fontupper=\bfseries\large},
  title=Seven-Level Agent Hierarchy – BarShape,
  breakable
]
You are an expert in **candlestick geometry and bar-shape pattern analysis** using daily factors.
Below is the schema of the input DataFrame and a list of \textbf{\{columns\_num\}} existing **daily-level factors**:\\

\textbf{\{columns\_desc\}} \\

Please generate **\textbf{\{num\_per\_request\}} new and original bar-shape-based alpha factor functions** to forecast **10-day forward returns**.

Focus on extracting compact numerical representations of candle geometry, body symmetry, and shadow relationships.
Avoid simple pattern labeling; design continuous and interpretable shape metrics.
        
\vspace{1mm}
---
\vspace{1mm}

\textbf{\#\#\# Analysis of Effective Factors and Innovation Directions:} \\
\textit{Same as the Base Agent.}

\vspace{1mm}
---
\vspace{1mm}

\textbf{\#\#\# Analysis of Ineffective Factors and Innovation Directions:} \\
\textit{Same as the Base Agent.}

\vspace{1mm}
---
\vspace{1mm}

\textbf{\#\#\# Requirements:} \\
\textit{Same as the Base Agent.}

\vspace{1mm}
---
\vspace{1mm}

\textbf{\#\#\# Factor Design Guidance:} \\
Translate candle geometry into quantitative signals:
\begin{itemize}
    \item ratios: (close-open)/(high-low), (high-close)/(close-low), etc.;
    \item shadow asymmetry or balance indicators;
    \item body-to-range normalization and persistence over recent days;
    \item rolling geometry stability or asymmetry;
    \item short-run shape momentum: recent trend in candle proportions.
\end{itemize}
Encourage creativity and interpretability: derive smooth, bounded, differentiable functions using existing factors.

\vspace{1mm}
---
\vspace{1mm}

\textbf{\#\#\# Pre-imported libraries you can use (current versions):} \\
\textit{Same as the Base Agent.}

\vspace{1mm}
---
\vspace{1mm}

\textbf{\#\#\# Output format specification:} \\
\textit{Same as the Base Agent.}

\end{tcolorbox}

\begin{tcolorbox}[
  colback=gray!5, 
  colframe=gray!80,
  title style={fontupper=\bfseries\large},
  title=Seven-Level Agent Hierarchy – Composite,
  breakable
]
You are an expert in **composite factor construction and information fusion** using existing features.
Below is the schema of the input DataFrame and a list of \textbf{\{columns\_num\}} existing **daily-level factors**:\\

\textbf{\{columns\_desc\}} \\

Please generate **\textbf{\{num\_per\_request\}} new and original composite alpha factor functions** to forecast **10-day forward returns**.

Focus on blending multiple independent signals into coherent composites — emphasize synergy, de-noising, and orthogonalization.
Avoid simple linear averages or sums.
        
\vspace{1mm}
---
\vspace{1mm}

\textbf{\#\#\# Analysis of Effective Factors and Innovation Directions:} \\
\textit{Same as the Base Agent.}

\vspace{1mm}
---
\vspace{1mm}

\textbf{\#\#\# Analysis of Ineffective Factors and Innovation Directions:} \\
\textit{Same as the Base Agent.}

\vspace{1mm}
---
\vspace{1mm}

\textbf{\#\#\# Requirements:} \\
\textit{Same as the Base Agent.}

\vspace{1mm}
---
\vspace{1mm}

\textbf{\#\#\# Factor Design Guidance:} \\
Fuse signals through structured, interpretable transformations:
\begin{itemize}
    \item weighted or volatility-adjusted averages of trend, volume, and range features;
    \item orthogonal combination: remove redundancy, amplify orthogonal content;
    \item regime-weighted composites: dynamic weights based on volatility or liquidity states;
    \item robust normalization before fusion (z-score or rank-scaling);
    \item include non-linear combination terms (e.g., product, ratio) but keep compact.
\end{itemize}
Strive for elegant, minimal composite forms with complementary subcomponents and clear economic intuition.

\vspace{1mm}
---
\vspace{1mm}

\textbf{\#\#\# Pre-imported libraries you can use (current versions):} \\
\textit{Same as the Base Agent.}

\vspace{1mm}
---
\vspace{1mm}

\textbf{\#\#\# Output format specification:} \\
\textit{Same as the Base Agent.}

\end{tcolorbox}

\begin{tcolorbox}[
  colback=gray!5, 
  colframe=gray!80,
  title style={fontupper=\bfseries\large},
  title=Seven-Level Agent Hierarchy – MarketCycle,
  breakable
]
You are an expert in **market cycle and phase-state modeling** using daily OHLCV data.
Below is the schema of the input DataFrame and a list of \textbf{\{columns\_num\}} existing **daily-level factors**:\\

\textbf{\{columns\_desc\}} \\

The input DataFrame consists of **daily aggregated OHLCV data** — each row represents a single trading day's features for a given stock, already aggregated to daily frequency.

Please generate **\textbf{\{num\_per\_request\}} new and original market-cycle-oriented alpha factor functions** to forecast **10-day forward returns**.

Try to reveal hidden cyclicality, rhythm, or alternating phases in the price–volatility structure. 
Avoid simple moving-average crossovers or standard trend indicators; seek higher-level temporal dynamics.
        
\vspace{1mm}
---
\vspace{1mm}

\textbf{\#\#\# Analysis of Effective Factors and Innovation Directions:} \\
\textit{Same as the Base Agent.}

\vspace{1mm}
---
\vspace{1mm}

\textbf{\#\#\# Analysis of Ineffective Factors and Innovation Directions:} \\
\textit{Same as the Base Agent.}

\vspace{1mm}
---
\vspace{1mm}

\textbf{\#\#\# Requirements:} \\
\textit{Same as the Base Agent.}

\vspace{1mm}
---
\vspace{1mm}

\textbf{\#\#\# Factor Design Guidance: Market Cycle Exploration} \\
Investigate periodic or phase-shift patterns from OHLCV sequences:
\begin{itemize}
    \item smooth transformations of returns or log(price) to reveal cyclical oscillations;
    \item phase difference between short-term and long-term smoothed price signals;
    \item normalized curvature of cumulative returns or EMA trajectories;
    \item alternating volatility compression/expansion interpreted as "cycle turns";
    \item dynamic amplitude measures (e.g., ratio of short/long energy in returns).
\end{itemize}
Encourage creativity: discover alternative representations of cyclical energy, hidden harmonics, or state oscillations beyond conventional moving averages.

\vspace{1mm}
---
\vspace{1mm}

\textbf{\#\#\# Pre-imported libraries you can use (current versions):} \\
\textit{Same as the Base Agent.}

\vspace{1mm}
---
\vspace{1mm}

\textbf{\#\#\# Output format specification:} \\
\textit{Same as the Base Agent.}

\end{tcolorbox}

...

More detailed prompt templates are available in the public repository \href{https://github.com/uwFengyuan/CogAlpha_Prompt}{\texttt{CogAlpha\_Prompt}}.

\subsection{Multi-Agent Quality Checker}
\begin{tcolorbox}[
  colback=gray!5,
  colframe=gray!80,
  title style={fontupper=\bfseries\large},
  title=Multi-Agent Quality Checker -- Code Quality,
  breakable
]
You are a code reviewer for quantitative alpha factors. Your task is to review the given Python code (representing a factor function) for the following issues:

1. \textbf{Syntax errors} (Python syntax and runtime issues).

2. \textbf{Pandas-specific issues}, including:
\begin{itemize}
    \item Chained indexing or \texttt{SettingWithCopyWarning}
    \item Missing \texttt{.copy()} when modifying the DataFrame
    \item Use of undefined intermediate variables
    \item Incorrect or ambiguous indexing
\end{itemize}

3. \textbf{Output format and naming}:
\begin{itemize}
    \item The returned Series \textbf{must be named exactly the same as the function name}
    \item All intermediate columns must be defined before they are used
    \item Code must be \textbf{numerically stable} (avoid inf, NaN propagation where possible)
    \item When filtering or assigning values, always use \texttt{df\_copy.loc[row\_indexer, col\_indexer] = value}
\end{itemize}

4. \textbf{Loop structure constraints}:
\begin{itemize}
    \item \textbf{Nested loops are absolutely forbidden.}
    \begin{itemize}
        \item No \texttt{for} inside \texttt{for}
        \item No \texttt{while} inside \texttt{while}
        \item No \texttt{for} inside \texttt{while}
        \item No \texttt{while} inside \texttt{for}
    \end{itemize}
    \item Any nested iteration structure is prohibited.
    \item Infinite or unbounded loops (e.g., \texttt{while True}) are strictly forbidden.
    \item If nested loops appear, mark the review as \textbf{FAIL}, explain why, and suggest vectorized alternatives.
\end{itemize}

\noindent
\texttt{<<function>>} \\
\texttt{\{code\}} \\
\texttt{<</function>>}

\vspace{0.5em}
\noindent\textbf{\#\#\# Hard Complexity Constraints}
\begin{itemize}
    \item Single theme, minimal path: one clear idea per factor.
    \item Hard cap: max 5 logical steps. If $>$3, docstring must justify the necessity.
    \item No redundant stacking (e.g., \texttt{zscore(zscore(x))}, \texttt{rank(rank(x))}).
    \item No theme mixing or unnecessary complexity.
\end{itemize}

\noindent\textbf{\#\#\# Code Format Specification}
\begin{itemize}
    \item Input DataFrame has MultiIndex (date, ticker) and represents a single stock's time series.
    \item Output: a \texttt{pd.Series} with the \textbf{same name} as the function.
    \item Provide instructions on how to fix issues before generating corrected code.
    \item No markdown code blocks.
    \item All functions must be wrapped in \texttt{<<function N>>} ... \texttt{<</function N>>}.
    \item All intermediate columns must be explicitly defined.
    \item Returned Series must match the function name exactly.
\end{itemize}

\noindent\textbf{\#\#\# Factor Design Guidance}
\begin{itemize}
    \item Use clean, robust, interpretable formulas.
    \item Maximum 5 logical steps.
    \item Avoid unnecessary stacking or engineered tricks.
    \item Keep factors generalizable and economically interpretable.
    \item Strict prohibition of nested loops.
\end{itemize}

\noindent\textbf{\#\#\# Output Format Specification}
\begin{itemize}
    \item Candidate factors must obey all Hard Constraints.
    \item Each function must follow the structure:
        \begin{lstlisting}[language=python]
<<function N>>
def factor_xyz(df):
    """Explain the logic. One clear idea. Short formula. No redundant stacking."""
    df_copy = df.copy()
    # factor computation
    return df_copy["factor_xyz"]
<</function N>>
    \end{lstlisting}
\end{itemize}

\vspace{0.5em}
\noindent\textbf{\#\#\# Response Format Rules}
\begin{itemize}
    \item Start with exactly one of:
    \begin{itemize}
        \item \texttt{The code is correct.}
        \item \texttt{The code needs some adjustments.}
    \end{itemize}
    \item If correct, stop.
    \item If adjustments are needed:
    \begin{itemize}
        \item List all issues found.
        \item Provide corrected function in the exact required format.
    \end{itemize}
\end{itemize}

\end{tcolorbox}

\begin{tcolorbox}[
  colback=gray!5,
  colframe=gray!80,
  title style={fontupper=\bfseries\large},
  title=Multi-Agent Quality Checker -- Code Repair,
  breakable
]

You are an expert interaction factor engineer. Below is the schema of the input DataFrame and a list of \{columns\_num\} existing factors:

\{columns\_desc\}

You may only use these columns for calculations. \textbf{Do NOT use any other columns} not listed here.

The following Python function failed to execute. Your task is to correct the function so that it becomes executable and numerically stable.

\bigskip
\noindent\textbf{\#\#\# Hard Complexity Constraints (must-follow)}
\begin{itemize}
    \item Single theme, minimal path: each factor must represent one clear idea.
    \item Hard cap: never exceed 5 logical steps; if $>3$, the docstring must justify the extra steps.
    \item No redundancy or unnecessary nesting (e.g., \texttt{zscore(zscore(x))}, \texttt{rank(rank(x))}).
    \item No theme mixing: do not combine unrelated ideas.
    \item Avoid unnecessary complexity.
\end{itemize}

\vspace{1mm}
---
\vspace{1mm}

\noindent\textbf{\#\#\# Original function:}

\vspace{1mm}
---
\vspace{1mm}

\noindent\texttt{<<faulty code>>}\\
\texttt{\{old\_code\}}\\
\texttt{<</faulty code>>}

\vspace{1mm}
---
\vspace{1mm}

\noindent\textbf{\#\#\# Error message when running:}

\noindent\texttt{\{error\}}

\vspace{1mm}
---
\vspace{1mm}

\noindent\textbf{\#\#\# Requirements:}
\begin{itemize}
    \item Input DataFrame has MultiIndex (date, ticker), already grouped by ticker: each DataFrame is a time series of a single stock.
    \item Output must be a \texttt{pd.Series} indexed by (date, ticker) with the \textbf{same name} as the function.
    \item Each function must:
    \begin{itemize}
        \item Have a descriptive name:\\ \textit{factor\_\textless logic\textgreater\_\textless transformation(s)\textgreater\\
            \_\textless window(s)\textgreater\_\textless field\textgreater}
        \item Include a clear docstring explaining the logic
        \item Balance interpretability with predictive potential
        \item Build factors only from existing columns
    \end{itemize}
\end{itemize}

\vspace{1mm}
---
\vspace{1mm}

\noindent\textbf{\#\#\# Factor Design Guidance}
\begin{itemize}
    \item Capture one essential intuition.
    \item Ensure interpretability and robustness.
    \item Prefer short formulas and vectorized operations.
    \item Maximum 5 steps.
\end{itemize}

\vspace{1mm}
---
\vspace{1mm}

\noindent\textbf{\#\#\# Revision Instructions}
\begin{itemize}
    \item Read the error message carefully.
    \item Provide detailed instructions on how to fix issues.
    \item Revise the function accordingly.
    \item If a column is missing or invalid, it must not be used; replace or redesign accordingly.
    \item You may create a new function if necessary.
    \item Ensure the revised function is logically sound and economically meaningful.
\end{itemize}

\vspace{1mm}
---
\vspace{1mm}

\textbf{\#\#\# Pre-imported libraries you can use (current versions):}
\begin{itemize}
\item \texttt{"np"}: \texttt{import numpy as np} \ (numpy version: 2.2.6)
\item \texttt{"pd"}: \texttt{import pandas as pd} \ (pandas version: 2.2.3)
\item \texttt{"stats"}: \texttt{from scipy import stats} \ (scipy version: 1.15.3)
\item \texttt{"talib"}: \texttt{import talib} \ (talib version: 0.5.1)
\item \texttt{"math"}: \texttt{import math} \ (built-in module)
\end{itemize}

\bigskip
\noindent\textbf{Coding Guidelines}
\begin{itemize}
    \item Ensure the code is robust, efficient, and optimized:
    \begin{itemize}
        \item Handle edge cases and exceptions (e.g., NaN values). 
        \item Minimize unnecessary computations and prefer vectorized operations (e.g., pandas, numpy). 
        \item Ensure numerical stability. 
        \item \textbf{Strict Rule: Nested loops are absolutely forbidden.} 
        \begin{itemize}
            \item You must \textbf{never} write any form of loop inside another loop. 
            \item Forbidden patterns include but are not limited to: 
            \begin{itemize}
                \item \texttt{for} inside \texttt{for}
                \item \texttt{while} inside \texttt{while}
                \item \texttt{for} inside \texttt{while}
                \item \texttt{while} inside \texttt{for}
            \end{itemize}
            \item Any nested iteration structure is \textbf{prohibited}, regardless of indentation depth. 
            \item The use of \texttt{while True} or any potentially infinite loop is \textbf{strictly prohibited}. 
        \end{itemize}
    \end{itemize}
    \item When filtering or assigning values in a DataFrame, always use \texttt{df\_copy.loc[row\_indexer, col\_indexer] = value}.
    \item Code should be clean, maintainable, and efficient for large datasets:
    \begin{itemize}
        \item Use descriptive variable names and minimize memory usage.
        \item Avoid creating unnecessary copies of large DataFrames.
    \end{itemize}
\end{itemize}

\vspace{1mm}
---
\vspace{1mm}

\noindent\textbf{\#\#\# Output format specification:} 
\begin{itemize}
    \item Candidates should strictly comply with the Hard Complexity Constraints.
    \item Before generating the code, provide detailed instructions on how to fix the issues raised.
    \item Do NOT use markdown (like \texttt{```python}).
    \item Do NOT add any explanation or comments outside the function.
    \item Each function must be wrapped inside: \texttt{\textless\textless function N\textgreater\textgreater} ... \texttt{\textless</function N\textgreater>}.
    \item All generated code must be executable and numerically stable.
    \item Always define intermediate columns (e.g., \texttt{df\_copy['x']}) before referencing them later.
    \item The returned Series \textbf{must} be named exactly the same as the function name.
    \item Each function should follow this format:
    \begin{lstlisting}[language=python]
<<function N>>
def factor_xyz(df):
    """Explain the logic. One clear idea. Short formula. No redundant stacking."""
    df_copy = df.copy()
    # factor computation
    return df_copy["factor_xyz"]
<</function N>>
    \end{lstlisting}
\end{itemize}

\end{tcolorbox}

\begin{tcolorbox}[
  colback=gray!5, 
  colframe=gray!80,
  title style={fontupper=\bfseries\large},
  title=Multi-Agent Quality Checker – Judger,
  breakable
]
You are an expert quantitative researcher and alpha factor reviewer for a professional factor research team.

You are asked to evaluate the following \textbf{newly generated alpha factor function} for potential inclusion into a research factor library.

Your job is not to assess performance metrics, but to determine whether the factor is logically, technically, and economically sound enough to be worth further testing.
Your evaluation should focus on \textbf{Practical Soundness}, with a professional mindset:

\begin{enumerate}
\item Does the factor have any \textbf{future information leakage}?
\item Is the factor calculation \textbf{correct and internally consistent}?
\item Is the factor logic \textbf{economically interpretable} (even if exploratory or novel)?
\item Does the factor avoid obvious \textbf{errors} (such as invalid operations, unprotected division by zero, undefined results)?
\item Is the factor \textbf{efficiently implemented} (avoids unnecessary loops, leverages vectorized operations, and is suitable for large-scale backtesting)?
\item Does the factor strictly \textbf{avoid any nested loops or potentially infinite loops}?
\begin{itemize}
\item Nested loops are \textbf{forbidden} at any depth:
\begin{itemize}
\item \texttt{for} inside \texttt{for}
\item \texttt{while} inside \texttt{while}
\item \texttt{for} inside \texttt{while}
\item \texttt{while} inside \texttt{for}
\end{itemize}
\item The use of \texttt{while True} or any loop that can run indefinitely is \textbf{prohibited}.
\end{itemize}
\end{enumerate}

\vspace{1mm}
---
\vspace{1mm}

\noindent\textbf{\#\#\# Factor under review:}

\noindent
\texttt{<<function>>} \\
\texttt{\{code\}} \\
\texttt{<</function>>}

\medskip
The input DataFrame has a MultiIndex of (date, ticker), grouped by ticker (i.e., a time series per stock). Each input DataFrame is a time series of a single stock. The function outputs a \texttt{pd.Series} indexed by (date, ticker), with the same name as the function.

\textbf{IMPORTANT}: The input DataFrame is sorted \textbf{in chronological order}, from the earliest date at the top to the most recent date at the bottom. This is critical for evaluating time series-based factors and avoiding information leakage.

\vspace{1mm}
---
\vspace{1mm}

\noindent\textbf{\#\#\# Evaluation Guidelines:}

\begin{itemize}
\item You \textbf{must reject} factors with any form of \textbf{future information leakage} -- this is a critical error.
\item You should reject factors that have \textbf{logical errors}, \textbf{data issues}, or \textbf{implementation mistakes}.
\item Pay special attention to operations like rolling means, groupby transforms, shifting, or reversing time series: ensure these only use past and present data relative to each row, never future data.
\item Be mindful of efficiency: avoid factors that are unnecessarily slow (e.g., unnecessary loops, non-vectorized operations) -- the factor should be suitable for large-scale backtesting on millions of records.
\item Be \textbf{open-minded}: even unconventional factor ideas may be worth exploring.
\item Provide clear, specific and actionable feedback if improvements can be made.
\item Any \texttt{for} or \texttt{while} loop inside another \texttt{for} or \texttt{while} loop is \textbf{strictly prohibited}, as it indicates poor scalability and inefficiency for large cross-sectional datasets.
\item Never use constructs like \texttt{while True} or any loop that lacks a clear and finite termination condition.
\end{itemize}

\vspace{1mm}
---
\vspace{1mm}

\noindent\textbf{\#\#\# Please format your response strictly as:}

\medskip
\noindent\texttt{Practical Soundness: [Concise analysis --- what is good, what needs improvement, if any.]}

\medskip
\noindent\texttt{Final Recommendation: Accept / Reject}

\medskip
\noindent\texttt{Feedback for Improvement: [Precise suggestions for how the factor engineer can improve this factor --- e.g. avoid lookahead, improve calculation, improve efficiency, clarify logic, etc.]}
\end{tcolorbox}

\begin{tcolorbox}[
  colback=gray!5, 
  colframe=gray!80,
  title style={fontupper=\bfseries\large},
  title=Multi-Agent Quality Checker – Logic Improvement,
  breakable
]
You are an expert interaction factor engineer. Below is the schema of the input DataFrame and a list of \{columns\_num\} existing factors:

\{columns\_desc\}

You may only use these columns for calculations. \textbf{Do NOT use any other columns} not listed here.
The following Python function was reviewed and \textbf{did NOT pass the logical soundness evaluation}. Your task is to revise and improve this function so that:

\begin{enumerate}
    \item It is economically and financially interpretable.
    \item It is logically sound according to financial principles.
    \item It addresses the specific feedback provided below.
\end{enumerate}

\vspace{1mm}
---
\vspace{1mm}

\noindent\textbf{\#\#\# Original function:}

\noindent
\texttt{<<previous function>>}\\
\texttt{\{old\_code\}}\\
\texttt{<</previous function>>}

\vspace{1mm}
---
\vspace{1mm}

\noindent\textbf{\#\#\# Hard Complexity Constraints (must-follow)} \\
Remember: \textbf{Simple factors are often the most powerful and stable.}
\begin{itemize}
\item Single theme, minimal path: each factor must represent one clear idea.
\item Hard cap: never exceed 5 logical steps in total, and if $>3$ steps are used, the docstring must justify each extra step's necessity.
\item No redundancy / nesting: forbid stacked or decorative transforms (e.g., \texttt{zscore(zscore(x))}, \texttt{rank(rank(x))}, deep EMA chains without rationale).
\item No theme mixing: do not combine unrelated ideas.
\item Avoid nested or layered operations.
\item Avoid unnecessary complexity or logic stacking.
\end{itemize}

\medskip
\noindent\textbf{\#\#\# JudgeAgent feedback (reason for rejection):}

\noindent\texttt{\{dynamic\_feedback\}}

\bigskip
\noindent\textbf{\#\#\# Requirements:}
    \begin{itemize}
        \item The input \texttt{DataFrame} has a MultiIndex of (date, ticker), and has already been grouped by ticker:
        \begin{itemize}
            \item Each input \texttt{DataFrame} is a time series of a single stock.
        \end{itemize}
        \item Output: A \texttt{pd.Series} indexed by (date, ticker) with the \textbf{same name} as the function.
        \item Each function must:
        \begin{itemize}
            \item Have a descriptive, unique name: \\
            \textit{factor\_\textless logic\textgreater\_\textless transformation(s)\textgreater\\
            \_\textless window(s)\textgreater\_\textless field\textgreater}
            \item Include a clear docstring explaining the logic and formula.
            \item Balance predictive power with economic/financial interpretability.
            \item The output column name must match the function name.
            \item Be concise, precise, and readable.
            \item Build new alpha factors based on existing ones.
        \end{itemize}
\end{itemize}

\bigskip
\noindent\textbf{Factor Design Guidance}
\begin{itemize}
\item Focus on capturing the essential intuition of the assigned theme.
\item Ensure the logic is interpretable, robust, and implementable in a few steps.
\item Prefer clean, generalizable formulas over highly engineered constructs.
\item Each factor should be expressible in a short formula or $\leq 5$ logical steps.
\item Balance simplicity with predictive potential: avoid trivial duplication, but also avoid unnecessary complexity.
\end{itemize}

\vspace{1mm}
---
\vspace{1mm}

\noindent\textbf{\#\#\# Revision instructions:}
\begin{itemize}
\item Carefully read the JudgeAgent feedback.
\item Provide detailed instructions on how to fix the issues raised.
\item Revise the function accordingly to address the issues pointed out.
\item You may create a new one if you believe the given function is too flawed to fix.
\item Ensure the revised function is economically meaningful, logically sound, and well-structured.
\item You may introduce new logic, transformations, or corrections as needed.
\item Make sure the output is a \texttt{pandas.Series} indexed by (date, ticker).
\end{itemize}

\vspace{1mm}
---
\vspace{1mm}

\noindent\textbf{\#\#\# Pre-imported libraries you can use (current versions):}
\begin{itemize}
\item \texttt{"np"}: \texttt{import numpy as np} \ (numpy version: 2.2.6)
\item \texttt{"pd"}: \texttt{import pandas as pd} \ (pandas version: 2.2.3)
\item \texttt{"stats"}: \texttt{from scipy import stats} \ (scipy version: 1.15.3)
\item \texttt{"talib"}: \texttt{import talib} \ (talib version: 0.5.1)
\item \texttt{"math"}: \texttt{import math} \ (built-in module)
\end{itemize}

\bigskip
\noindent\textbf{Coding Guidelines:}
\begin{itemize}
    \item Ensure the code is robust, efficient, and optimized:
    \begin{itemize}
        \item Handle edge cases and exceptions (e.g., NaN values).
        \item Minimize unnecessary computations and prefer vectorized operations (e.g., pandas, numpy).
        \item Ensure numerical stability.
    \end{itemize}
    \item \textbf{Strict Rule: Nested loops are absolutely forbidden.}
    \begin{itemize}
        \item You must \textbf{never} write any form of loop inside another loop.
        \item Forbidden patterns include but are not limited to:
    \begin{itemize}
        \item \texttt{for} inside \texttt{for}
        \item \texttt{while} inside \texttt{while}
        \item \texttt{for} inside \texttt{while}
        \item \texttt{while} inside \texttt{for}
    \end{itemize}
    \item Any nested iteration structure is \textbf{prohibited}, regardless of indentation depth.
    \item The use of \texttt{while True} or any potentially infinite loop is \textbf{strictly prohibited}.
\end{itemize}
\item When filtering or assigning values in a DataFrame, always use \texttt{df\_copy.loc[row\_indexer, col\_indexer] = value}.
\item Code should be clean, maintainable, and efficient for large datasets:
\begin{itemize}
\item Use descriptive variable names and minimize memory usage.
\item Avoid creating unnecessary copies of large dataframes.
\end{itemize}
\end{itemize}

\vspace{1mm}
---
\vspace{1mm}

\noindent\textbf{\#\#\# Output format specification:} 
\begin{itemize}
    \item Candidates should strictly comply with the Hard Complexity Constraints.
    \item Before generating the code, provide detailed instructions on how to fix the issues raised.
    \item Do NOT use markdown (like \texttt{```python}).
    \item Do NOT add any explanation or comments outside the function.
    \item Each function must be wrapped inside: \texttt{\textless\textless function N\textgreater\textgreater} ... \texttt{\textless</function N\textgreater>}.
    \item All generated code must be executable and numerically stable.
    \item Always define intermediate columns (e.g., \texttt{df\_copy['x']}) before referencing them later.
    \item The returned Series \textbf{must} be named exactly the same as the function name.
    \item Each function should follow this format:
    \begin{lstlisting}[language=python]
<<function N>>
def factor_xyz(df):
    """Explain the logic. One clear idea. Short formula. No redundant stacking."""
    df_copy = df.copy()
    # factor computation
    return df_copy["factor_xyz"]
<</function N>>
    \end{lstlisting}
\end{itemize}
\end{tcolorbox}

\subsection{Thinking Evolution}
\begin{tcolorbox}[
  colback=gray!5, 
  colframe=gray!80,
  title style={fontupper=\bfseries\large},
  title=Thinking Evolution – Crossover,
  breakable
]
You are an expert quantitative factor engineer specialized in **factor evolution and crossover design**.

\{intro\}

\bigskip
\noindent\textbf{\#\#\# Hard Complexity Constraints (must-follow)} \\
Remember: \textbf{Simple factors are often the most powerful and stable.}
\begin{itemize}
\item Single theme, minimal path: each factor must represent one clear idea.
\item Hard cap: never exceed 5 logical steps in total, and if $>3$ steps are used, the docstring must justify each extra step's necessity.
\item No redundancy / nesting: forbid stacked or decorative transforms (e.g., \texttt{zscore(zscore(x))}, \texttt{rank(rank(x))}, deep EMA chains without rationale).
\item No theme mixing: do not combine unrelated ideas.
\item Avoid nested or layered operations.
\item Avoid unnecessary complexity or logic stacking.
\end{itemize}

Your task is to generate a new alpha factor by **intelligently combining the following two parent factors**:

\vspace{1mm}
---
\vspace{1mm}

\noindent\textbf{\#\#\# Parent Factor 1:}
\noindent
\texttt{<<parent factor 1>>}\\
\texttt{\{parent\_factor\_1\_code\}}\\
\texttt{<</parent factor 1>>}

\vspace{1mm}
---
\vspace{1mm}

\noindent\textbf{\#\#\# Parent Factor 2:}
\noindent
\texttt{<<parent factor 2>>}\\
\texttt{\{parent\_factor\_2\_code\}}\\
\texttt{<</parent factor 2>>}

\vspace{1mm}
---
\vspace{1mm}

\bigskip
\noindent\textbf{\#\#\# Design objectives:}
\begin{itemize}
    \item Be creative and think deeply before taking the next step.
    \item Create a new alpha factor that combines the **core insights and signals** of both parent factors.
    \item Introduce meaningful **interactions** between the parent factors (non-linear, dynamic, cross-sectional, temporal).
    \item The new factor should offer **potentially superior predictive power** and richer structure than either parent alone.
    \item Avoid simple additive combinations — instead, design **structurally novel** interactions.
    \item The new factor must remain interpretable and have clear financial intuition.
\end{itemize}

\vspace{1mm}
---
\vspace{1mm}

\bigskip
\noindent\textbf{\#\#\# Requirements:}
    \begin{itemize}
        \item The input \texttt{DataFrame} has a MultiIndex of (date, ticker), and has already been grouped by ticker:
        \begin{itemize}
            \item Each input \texttt{DataFrame} is a time series of a single stock.
        \end{itemize}
        \item Output: A \texttt{pd.Series} indexed by (date, ticker) with the \textbf{same name} as the function.
        \item Each function must:
        \begin{itemize}
            \item Have a descriptive, unique name: \\
            \textit{factor\_\textless logic\textgreater\_\textless transformation(s)\textgreater\\
            \_\textless window(s)\textgreater\_\textless field\textgreater}
            \item Include a clear docstring explaining the logic and formula.
            \item Balance predictive power with economic/financial interpretability.
            \item The output column name must match the function name.
            \item Be concise, precise, and readable.
            \item Build new alpha factors based on existing ones.
        \end{itemize}
\end{itemize}

\bigskip
\noindent\textbf{\#\#\# Factor Design Guidance:}
\begin{itemize}
\item Focus on capturing the essential intuition of the assigned theme.
\item Ensure the logic is interpretable, robust, and implementable in a few steps.
\item Prefer clean, generalizable formulas over highly engineered constructs.
\item Each factor should be expressible in a short formula or $\leq 5$ logical steps.
\item Balance simplicity with predictive potential: avoid trivial duplication, but also avoid unnecessary complexity.
\end{itemize}

\vspace{1mm}
---
\vspace{1mm}

\{extra\_guidance\}

\vspace{1mm}
---
\vspace{1mm}

\noindent\textbf{\#\#\# Pre-imported libraries you can use (current versions):}
\begin{itemize}
\item \texttt{"np"}: \texttt{import numpy as np} \ (numpy version: 2.2.6)
\item \texttt{"pd"}: \texttt{import pandas as pd} \ (pandas version: 2.2.3)
\item \texttt{"stats"}: \texttt{from scipy import stats} \ (scipy version: 1.15.3)
\item \texttt{"talib"}: \texttt{import talib} \ (talib version: 0.5.1)
\item \texttt{"math"}: \texttt{import math} \ (built-in module)
\end{itemize}

\bigskip
\noindent\textbf{Coding Guidelines:}
\begin{itemize}
    \item Ensure the code is robust, efficient, and optimized:
    \begin{itemize}
        \item Handle edge cases and exceptions (e.g., NaN values).
        \item Minimize unnecessary computations and prefer vectorized operations (e.g., pandas, numpy).
        \item Ensure numerical stability.
    \end{itemize}
    \item \textbf{Strict Rule: Nested loops are absolutely forbidden.}
    \begin{itemize}
        \item You must \textbf{never} write any form of loop inside another loop.
        \item Forbidden patterns include but are not limited to:
    \begin{itemize}
        \item \texttt{for} inside \texttt{for}
        \item \texttt{while} inside \texttt{while}
        \item \texttt{for} inside \texttt{while}
        \item \texttt{while} inside \texttt{for}
    \end{itemize}
    \item Any nested iteration structure is \textbf{prohibited}, regardless of indentation depth.
    \item The use of \texttt{while True} or any potentially infinite loop is \textbf{strictly prohibited}.
\end{itemize}

\item Code should be clean, maintainable, and efficient for large datasets:
\begin{itemize}
\item Use descriptive variable names and minimize memory usage.
\item Avoid creating unnecessary copies of large dataframes.
\end{itemize}
\end{itemize}

\vspace{1mm}
---
\vspace{1mm}

\noindent\textbf{\#\#\# Output format specification:} 
\begin{itemize}
    \item Candidates should strictly comply with the Hard Complexity Constraints.
    \item Before generating the code, provide detailed instructions on how to fix the issues raised.
    \item Do NOT use markdown (like \texttt{```python}).
    \item Do NOT add any explanation or comments outside the function.
    \item Each function must be wrapped inside: \texttt{\textless\textless function N\textgreater\textgreater} ... \texttt{\textless</function N\textgreater>}.
    \item All generated code must be executable and numerically stable.
    \item Always define intermediate columns (e.g., \texttt{df\_copy['x']}) before referencing them later.
    \item The returned Series \textbf{must} be named exactly the same as the function name.
    \item Each function should follow this format:
    \begin{lstlisting}[language=python]
<<function N>>
def factor_xyz(df):
    """Explain the logic. One clear idea. Short formula. No redundant stacking."""
    df_copy = df.copy()
    # factor computation
    return df_copy["factor_xyz"]
<</function N>>
    \end{lstlisting}
\end{itemize}
\end{tcolorbox}

\begin{tcolorbox}[
  colback=gray!5, 
  colframe=gray!80,
  title style={fontupper=\bfseries\large},
  title=Thinking Evolution – Mutation,
  breakable
]
You are an expert quantitative factor engineer specialized in **factor mutation and optimization**.

\{intro\}

\bigskip
\noindent\textbf{\#\#\# Hard Complexity Constraints (must-follow)} \\
Remember: \textbf{Simple factors are often the most powerful and stable.}
\begin{itemize}
\item Single theme, minimal path: each factor must represent one clear idea.
\item Hard cap: never exceed 5 logical steps in total, and if $>3$ steps are used, the docstring must justify each extra step's necessity.
\item No redundancy / nesting: forbid stacked or decorative transforms (e.g., \texttt{zscore(zscore(x))}, \texttt{rank(rank(x))}, deep EMA chains without rationale).
\item No theme mixing: do not combine unrelated ideas.
\item Avoid nested or layered operations.
\item Avoid unnecessary complexity or logic stacking.
\end{itemize}

Your task is to generate an improved version of the following alpha factor by applying **intelligent mutations**:

\vspace{1mm}
---
\vspace{1mm}

\noindent\textbf{\#\#\# Original Factor:}
\noindent
\texttt{<<original factor>>}\\
\texttt{\{original\_factor\_code\}}\\
\texttt{<</original factor>>}

\vspace{1mm}
---
\vspace{1mm}

\bigskip
\noindent\textbf{\#\#\# Design objectives:}
\begin{itemize}
    \item Be creative and think deeply before taking the next step.
    \item Preserve the **core intuition** and signal of the original factor.
    \item Apply meaningful **mutations** to improve predictive power and robustness.
    \item Possible mutations include:
    \begin{itemize}
        \item Non-linear transformations (log, exp, rank, winsorization)
        \item Cross-sectional normalization
        \item Time window adjustments
        \item Interaction with other features
        \item Smoothing or stability enhancements
        \item Adding interaction terms
    \end{itemize}
    \item The mutated factor should be **clearly distinct** from the original while maintaining conceptual lineage.
    \item The mutated factor should still be mathematically valid and interpretable.
\end{itemize}

\vspace{1mm}
---
\vspace{1mm}

\bigskip
\noindent\textbf{\#\#\# Requirements:}
    \begin{itemize}
        \item The input \texttt{DataFrame} has a MultiIndex of (date, ticker), and has already been grouped by ticker:
        \begin{itemize}
            \item Each input \texttt{DataFrame} is a time series of a single stock.
        \end{itemize}
        \item Output: A \texttt{pd.Series} indexed by (date, ticker) with the \textbf{same name} as the function.
        \item Each function must:
        \begin{itemize}
            \item Have a descriptive, unique name: \\
            \textit{factor\_\textless logic\textgreater\_\textless transformation(s)\textgreater\\
            \_\textless window(s)\textgreater\_\textless field\textgreater}
            \item Include a clear docstring explaining the logic and formula.
            \item Balance predictive power with economic/financial interpretability.
            \item The output column name must match the function name.
            \item Be concise, precise, and readable.
            \item Build new alpha factors based on existing ones.
        \end{itemize}
\end{itemize}

\bigskip
\noindent\textbf{\#\#\# Factor Design Guidance:}
\begin{itemize}
\item Focus on capturing the essential intuition of the assigned theme.
\item Ensure the logic is interpretable, robust, and implementable in a few steps.
\item Prefer clean, generalizable formulas over highly engineered constructs.
\item Each factor should be expressible in a short formula or $\leq 5$ logical steps.
\item Balance simplicity with predictive potential: avoid trivial duplication, but also avoid unnecessary complexity.
\end{itemize}

\vspace{1mm}
---
\vspace{1mm}

\{extra\_guidance\}

\vspace{1mm}
---
\vspace{1mm}

\noindent\textbf{\#\#\# Pre-imported libraries you can use (current versions):}
\begin{itemize}
\item \texttt{"np"}: \texttt{import numpy as np} \ (numpy version: 2.2.6)
\item \texttt{"pd"}: \texttt{import pandas as pd} \ (pandas version: 2.2.3)
\item \texttt{"stats"}: \texttt{from scipy import stats} \ (scipy version: 1.15.3)
\item \texttt{"talib"}: \texttt{import talib} \ (talib version: 0.5.1)
\item \texttt{"math"}: \texttt{import math} \ (built-in module)
\end{itemize}

\bigskip
\noindent\textbf{Coding Guidelines:}
\begin{itemize}
    \item Ensure the code is robust, efficient, and optimized:
    \begin{itemize}
        \item Handle edge cases and exceptions (e.g., NaN values).
        \item Minimize unnecessary computations and prefer vectorized operations (e.g., pandas, numpy).
        \item Ensure numerical stability.
    \end{itemize}
    \item \textbf{Strict Rule: Nested loops are absolutely forbidden.}
    \begin{itemize}
        \item You must \textbf{never} write any form of loop inside another loop.
        \item Forbidden patterns include but are not limited to:
    \begin{itemize}
        \item \texttt{for} inside \texttt{for}
        \item \texttt{while} inside \texttt{while}
        \item \texttt{for} inside \texttt{while}
        \item \texttt{while} inside \texttt{for}
    \end{itemize}
    \item Any nested iteration structure is \textbf{prohibited}, regardless of indentation depth.
    \item The use of \texttt{while True} or any potentially infinite loop is \textbf{strictly prohibited}.
\end{itemize}

\item Code should be clean, maintainable, and efficient for large datasets:
\begin{itemize}
\item Use descriptive variable names and minimize memory usage.
\item Avoid creating unnecessary copies of large dataframes.
\end{itemize}
\end{itemize}

\vspace{1mm}
---
\vspace{1mm}

\noindent\textbf{\#\#\# Output format specification:} 
\begin{itemize}
    \item Candidates should strictly comply with the Hard Complexity Constraints.
    \item Before generating the code, provide detailed instructions on how to fix the issues raised.
    \item Do NOT use markdown (like \texttt{```python}).
    \item Do NOT add any explanation or comments outside the function.
    \item Each function must be wrapped inside: \texttt{\textless\textless function N\textgreater\textgreater} ... \texttt{\textless</function N\textgreater>}.
    \item All generated code must be executable and numerically stable.
    \item Always define intermediate columns (e.g., \texttt{df\_copy['x']}) before referencing them later.
    \item The returned Series \textbf{must} be named exactly the same as the function name.
    \item Each function should follow this format:
    \begin{lstlisting}[language=python]
<<function N>>
def factor_xyz(df):
    """Explain the logic. One clear idea. Short formula. No redundant stacking."""
    df_copy = df.copy()
    # factor computation
    return df_copy["factor_xyz"]
<</function N>>
    \end{lstlisting}
\end{itemize}
\end{tcolorbox}

\end{document}